\documentclass[10pt,twocolumn,letterpaper]{article}

\usepackage{iccv}              

\usepackage{amsmath,amsfonts}
\usepackage{algorithmic}
\usepackage{algorithm}
\usepackage{array}
\usepackage[accsupp]{axessibility}
\usepackage{textcomp}
\usepackage{stfloats}
\usepackage{url}
\usepackage{verbatim}

\usepackage[utf8]{inputenc} 
\usepackage[T1]{fontenc}    
\usepackage{url}            
\usepackage{booktabs}       
\usepackage{amsfonts}       
\usepackage{nicefrac}       
\usepackage{microtype}      
\usepackage{xcolor}         

\usepackage{xspace}
\usepackage{amssymb}
\usepackage{graphicx}
\usepackage{enumitem}
\usepackage{bm}
\usepackage{bbding}
\usepackage{pifont}

\usepackage{subcaption}
\usepackage{colortbl}
\usepackage{wrapfig}
\usepackage[font=small]{caption}

\graphicspath{{imgs/}}
\setlist[itemize]{itemsep=0pt,topsep=0pt,partopsep=0pt,parsep=0pt,leftmargin=4mm}
\usepackage{array}
\newcolumntype{C}{>{$}c<{$}} 

\definecolor{myorange}{RGB}{255, 127, 102}
\definecolor{mygreen}{RGB}{120, 210, 170}
\definecolor{myblue}{RGB}{0, 102, 204}
\definecolor{lightblue}{RGB}{232, 244, 248}
\definecolor{myred}{RGB}{220, 60, 60}
\definecolor{mypurple}{RGB}{160, 130, 210}
\definecolor{mypink}{RGB}{220, 100, 120}

\definecolor{greenyellow}{RGB}{197,228,10}
\newcommand{\first}[1]{\textbf{#1}}

\newcommand{\second}[1]{\underline{#1}}

  \def\cL{{\mathcal L}} \def\cR{{\mathcal R}} \def\cX{{\mathcal X}}

\newcommand{\impsq}{\Psi}

\newcommand{\pointset}{\Omega}

\definecolor{iccvblue}{rgb}{0.21,0.49,0.74}
\usepackage[pagebackref,breaklinks,colorlinks,allcolors=iccvblue]{hyperref}

\def\paperID{9186} 
\def\confName{ICCV}
\def\confYear{2025}

\title{Self-supervised Learning of Hybrid Part-aware 3D Representations \\ of 2D Gaussians and Superquadrics}

\author{
Zhirui Gao  \quad 
Renjiao Yi$^{\dagger}$  \quad
Yuhang Huang  \quad
 Wei Chen  \quad
Chenyang Zhu  \quad
Kai Xu$^{\dagger}$ \\
 National University of Defense Technology\\
{\href{https://zhirui-gao.github.io/PartGS}{zhirui-gao.github.io/PartGS}}
}

\begin{document}
\maketitle
\let\thefootnote\relax\footnotetext{$^{\dagger}$ Corresponding authors}
\begin{abstract}
Low-level 3D representations, such as point clouds, meshes, NeRFs and 3D Gaussians, are commonly used for modeling 3D objects and scenes. However, cognitive studies indicate that human perception operates at higher levels and interprets 3D environments by decomposing them into meaningful structural parts, rather than low-level elements like points or voxels. Structured geometric decomposition enhances scene interpretability and facilitates downstream tasks requiring component-level manipulation. In this work, we introduce \textit{\textbf{PartGS}}, a self-supervised \textbf{part}-aware reconstruction framework that integrates 2D \textbf{G}aussians and \textbf{s}uperquadrics to parse objects and scenes into an interpretable decomposition, leveraging multi-view image inputs to uncover 3D structural information. Our method jointly optimizes superquadric meshes and Gaussians by coupling their parameters within a hybrid representation. On one hand, superquadrics enable the representation of a wide range of shape primitives, facilitating flexible and meaningful decompositions. On the other hand, 2D Gaussians capture detailed texture and geometric details, ensuring high-fidelity appearance and geometry reconstruction. Operating in a self-supervised manner, our approach demonstrates superior performance compared to state-of-the-art methods across extensive experiments on the DTU, ShapeNet, and real-world datasets.
\end{abstract}    
\section{Introduction}

\begin{figure}
  \centering
\includegraphics[width=\linewidth]{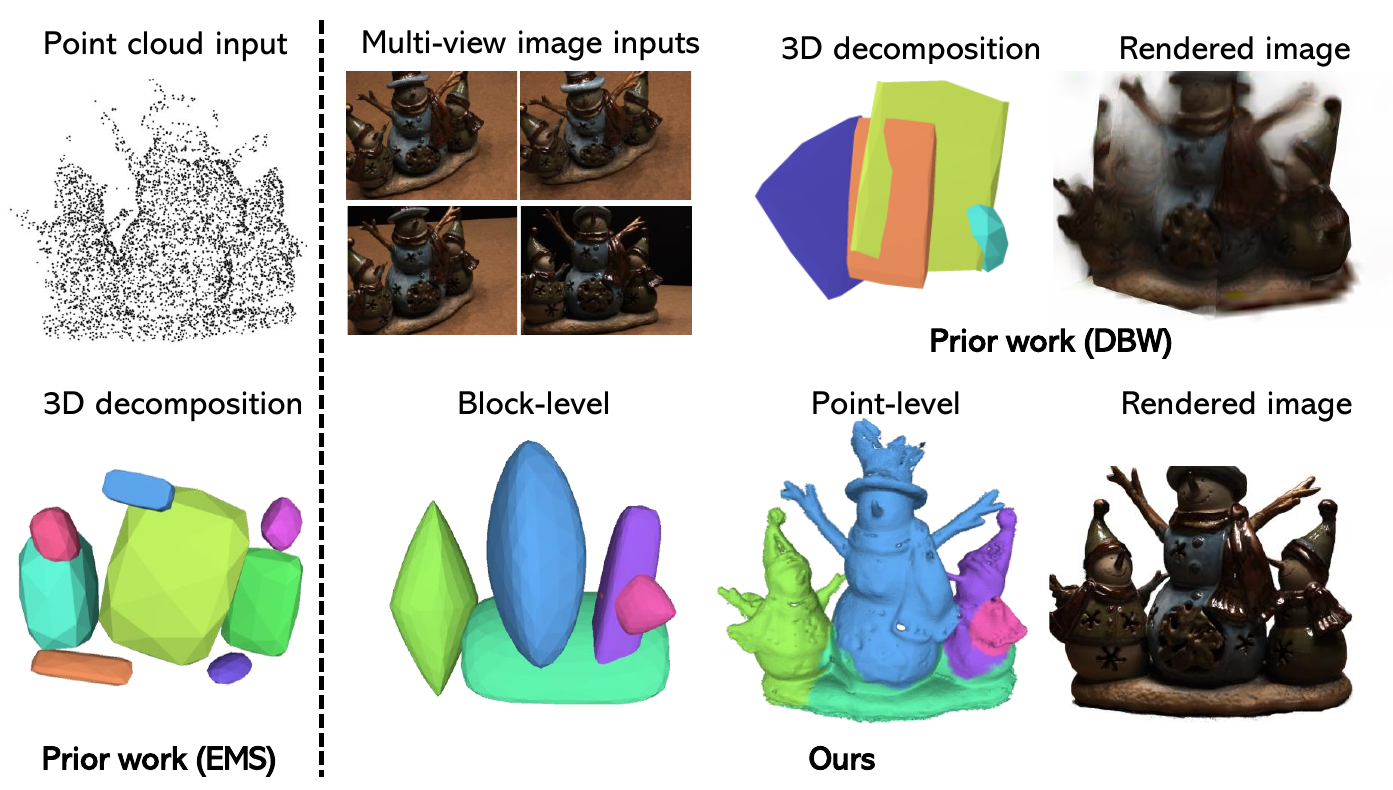}
  \caption{Prior works~\cite{liu2022robust, monnier2023dbw} model the scene using primitives, while
the proposed method can further model precise geometry details and textures. Here, EMS~\cite{liu2022robust} takes point cloud inputs and reconstructs
non-textured primitives.}
  \label{fig:teaser}
  \vspace{-8px}
\end{figure}

\label{sec:intro}

3D reconstruction from multi-view images is a long-standing challenge in 3D vision and graphics~\cite{wang2021neus,gao2025generic,gao2024fdc}. Most reconstructed scenes are in low-level representations such as point clouds, voxels, or meshes. However, humans tend to understand 3D scenes as reasonable parts~\cite{Mitra_Wand_Zhang_Cohen-Or_Kim_Huang_2013}. For instance, when observing a scene, we naturally construct high-level structural information, such as scene graphs, instead of focusing on low-level details like point clouds or voxels.   Motivated by it, we propose a part-aware reconstruction framework that decomposes objects or scenes into meaningful shapes or parts, facilitating tasks such as physical simulation, editing,  content generation and understanding~\cite{Gu_2025_CVPR,li2017grass,Wang_2025_CVPR, wang2025geollava, wang2025xlrs}.

Several prior works~\cite{zhu2020adacoseg, Niu_Li_Xu_2018,  li2019supervised, liu2022robust, 
loiseau2023learnable, wang2020pie} have explored part-aware reconstruction or 3D decomposition. However, most of them rely heavily on 3D supervision and often struggle to retain fine-grained geometric details, limiting their practicality in real-world scenarios.  Recent advances \cite{tertikas2023partnerf,yu2024dpa} have focused on extending the neural radiance field framework for part-aware reconstruction, building upon its remarkable success in 3D reconstruction from multi-view images.  For example, PartNeRF~\cite{tertikas2023partnerf} models objects using multiple neural radiance fields. Yet, the intricate composition of implicit fields complicates the learning process, leading to suboptimal rendering quality and inefficient decomposition.  Recently, DBW \cite{monnier2023dbw} proposes a novel approach that decomposes scenes into block-based representations using superquadric primitives~\cite{barr1981superquadrics}, optimizing both their parameters and UV texture maps through rendering loss minimization. While this method demonstrates effective scene decomposition into coherent components, it faces challenges in achieving accurate part-aware reconstruction of both geometry and appearance, as illustrated in Fig.~\ref{fig:teaser}.

To address these limitations, we propose PartGS, a self-supervised part-aware hybrid representation that integrates 2D Gaussians~\cite{Huang2DGS2024} and superquadrics~\cite{barr1981superquadrics}, to achieve both high-quality texture reconstruction and geometrically accurate part decomposition. There are previous approaches~\cite{guedon2023sugar, gao2024mesh,waczynska2024games} combining mesh reconstruction and Gaussian splatting, where they first reconstruct the mesh and then attach Gaussians to it. In contrast to their primary focus on acquiring the representations, the proposed approach is to achieve part-aware decomposition reconstruction.  Our method establishes a coupled optimization framework that simultaneously learns superquadric meshes and Gaussians through parameter sharing, with each part constrained to a single superquadric. The differentiable rendering of Gaussians drives this hybrid representation, leveraging the inherent convexity of superquadrics for qualitative 3D decomposition. The self-supervised training of part-aware reconstruction is built upon the assumption that each part of most objects should be a basic shape and can be represented by a superquadric. Meanwhile, Gaussian splatting, renowned for its superior rendering quality and efficient training, is incorporated to capture intricate texture details, speeding up reconstruction and improving rendering quality.

In the hybrid representation, 2D Gaussians are distributed on superquadric surfaces to form structured blocks. The pose and shape of Gaussians within each block are determined by their corresponding superquadrics rather than being optimized independently. The parameter space encompasses global controls for superquadric properties (shape, pose, and opacity) and local spherical harmonic coefficients for individual Gaussians. Compared to standard Gaussian splatting~\cite{kerbl20233d, Huang2DGS2024}, which populates the occupancy volume with independently parameterized Gaussians, the coupled representation is more compact and efficient.

During training, the parameters are optimized through rendering loss without additional supervision. However, we notice that image rendering loss often leads to local minima in superquadric shape optimization. To tackle this, we introduce several regularizers to maintain global consistency between the 3D representation and the input 2D information. This strategic implementation allows us to segment parts fully self-supervised, achieving block-level reconstruction. Finally, to better capture irregular shapes, we implement a point-level refinement step that frees 2D Gaussians to deviate from the surface, thereby enhancing geometric fidelity. Extensive experiments show that PartGS, at block-level and point-level,  achieves {33.3\%}  and {75.9\%}  improvements in reconstruction accuracy, {3.18} and {16.13} increases in PSNR, and {4X} and {3X} speedups compared to the state-of-the-art baseline. 
Our contributions are summarized as follows:
    \begin{itemize}
        \item We introduce a novel hybrid representation for part-aware 3D reconstruction, combining the strengths of superquadrics and Gaussian splatting to achieve reasonable decomposition and high-quality rendering.
        \item We propose several novel regularizers to enforce consistency between 3D decomposition and 2D observations, enabling self-supervised part decomposition. 

        \item Compared to prior works, the method takes one step forward, by reconstructing both the block-level and point-level part-aware reconstructions, preserving both part segmentation and reconstruction precision. 
        
    \end{itemize}
\begin{figure*}[t]
  \centering
  \includegraphics[width=0.95\textwidth]{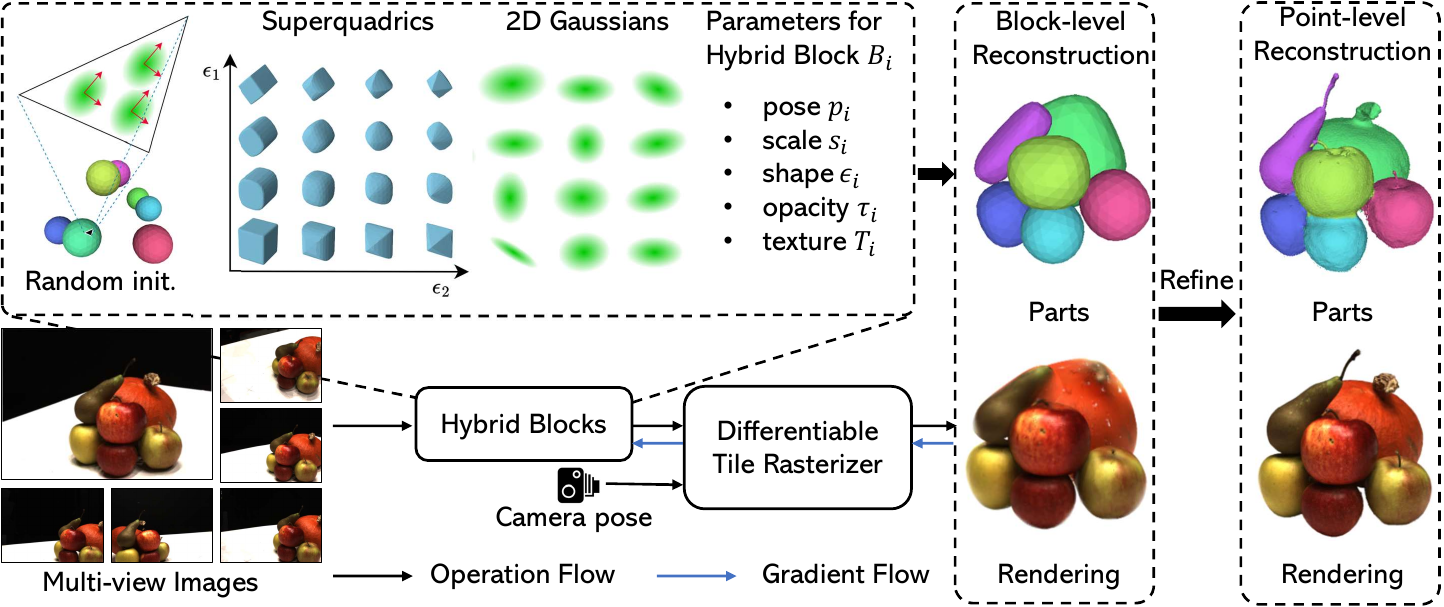}
  \caption{\textbf{Overview of our pipeline.} PartGS takes multi-view images to learn a parametric hybrid representation of superquadrics and 2D Gaussians. It initializes from random superquadrics and is gradually optimized during training to obtain a block-level reconstruction. Then, we free the constraints of Gaussians to model detailed geometry, achieving point-level reconstruction.}
  \label{fig:pip}
   \vspace{-6px}
\end{figure*}

\section{Related Work}

\subsection{Shape Decomposition and 
 Abstraction}

Structured shape representation learning decomposes objects or scenes into coherent geometric primitives, facilitating shape understanding and generation~\cite{STAR-TM,guan2020fame,gao2024learning, chen2023dae,sun2022semi}. Early approaches like Blocks World~\cite{roberts1963machine} and Generalized Cylinders~\cite{binford1971visual} emphasized compact representations. Modern methods typically process 3D inputs (point clouds, meshes, voxels) by decomposing them into primitive ensembles, including cuboids~\cite{tulsiani2017learning,ramamonjisoa2022monteboxfinder}, superquadrics~\cite{paschalidou2019superquadrics,paschalidou2020learning,wu2022primitivebased,liu2022robust}, and convex shapes~\cite{Chen_Tagliasacchi_Zhang_2020,Deng_Genova_Yazdani_Bouaziz_Hinton_Tagliasacchi_2020}. For instance, MonteBoxFinder~\cite{ramamonjisoa2022monteboxfinder} integrates clustering~\cite{fischler1981random,10486880,CCGC}, cuboid representations, and Monte Carlo Tree Search, while EMS~\cite{liu2022robust} adopts a probabilistic approach for superquadric recovery. However, these methods are limited to coarse shape representations. Recent advances in shape abstraction~\cite{gao2019sdm, paschalidou2021neural,hui2022neural,shuai2023dpf,liu2023deformer} enable more detailed shape representations through flexible primitive deformation.


Some studies~\cite{tertikas2023partnerf, monnier2023dbw, yu2024dpa, Alaniz_2023_ICCV, gao2025curveawaregaussiansplatting3d, zhou2025monomobilityzeroshot3dmobility} attempt to create structure-aware 3D representations directly from images. PartNeRF~\cite{tertikas2023partnerf} introduces ellipsoid representations within NeRFs, its reliance on multiple implicit fields results in inefficient 3D decomposition. ISCO~\cite{Alaniz_2023_ICCV} and DBW~\cite{monnier2023dbw} use 3D superquadrics for shape decomposition, which enables more meaningful structural separation. However, their simple shape parameters struggle to capture complex geometries, leading to poor geometry and appearance reconstruction. DPA-Net~\cite{yu2024dpa} has advanced 3D shape abstraction from sparse views but generates redundant parts and struggles with realistic texture rendering. 
A concurrent work, GaussianBlock~\cite{jiang2024gaussianblock}, employs SAM~\cite{kirillov2023segany} to guide superquadric splitting and fusion for 3D decomposition, yet its computational efficiency remains limited, typically requiring several hours per scene. In contrast, our approach accomplishes self-supervised~\cite{ICL-SSL, wan2024contrastive} part-aware scene reconstruction through an efficient hybrid representation that simultaneously maintains geometric fidelity and photorealistic rendering quality.

\subsection{Mesh-based Gaussian Splatting}  
Gaussian splitting (GS)~\cite{kerbl20233d} has been rapidly adopted in multiple fields due to its remarkable rendering capability.  Several studies~\cite{guedon2023sugar,chen2023neusg,gao2024mesh,waczynska2024games} attempt to align Gaussians with mesh surfaces for easier editing and animation. Among them, SuGaR~\cite{guedon2023sugar} uses flat 3D Gaussians to enforce the alignment with the scene surface during optimization, minimizing the difference between the signed distance functions (SDF) of the desired Gaussians and actual Gaussians. GaMeS~\cite{waczynska2024games} introduces a hybrid representation of Gaussians and mesh, where Gaussians are attached to triangular facets of the mesh. Similarly,  Gao~\etal~\cite{gao2024mesh} proposes a mesh-based GS to achieve large-scale deformation effects on objects.  Recently, 2DGS~\cite{Huang2DGS2024} proposes 2D Gaussians for surface modeling and significantly enhances the geometric quality. Inspired by them, we propose a representation that combines 2D Gaussians with superquadrics.   A key distinction between our method and previous mesh-based GS approaches is that these methods require first reconstructing the scene's mesh and then binding Gaussians to the mesh surface, which results in a non-continuous mesh. In contrast, our method directly optimizes the mesh through a rendering loss, enabling part-aware mesh reconstruction.


\section{Method}

Given a set of calibrated multi-view images and foreground masks, we aim to learn part-aware 3D representations, a meaningful decomposition of both geometry and appearance. Our approach adopts a two-stage optimization strategy: first, decomposing the object into basic shapes using a mixture of Gaussians and superquadrics at the block-level, followed by refining the decomposition at the point-level to achieve precise geometry. In Sec~\ref{sec:hybrid}, we parameterize the hybrid representation. Sec.~\ref{sec:block-opt} then elaborates on leveraging this representation, enhanced by novel regularizers and an adaptive control strategy, to achieve self-supervised block-level decomposition. Finally, Sec.~\ref{sec:point-opt} presents the process for obtaining detailed part-aware results.

\subsection{Parametrizing the Hybrid Representation}\label{sec:hybrid}

As shown in Fig.~\ref{fig:pip}, to leverage the strengths of both, we attach Gaussians to the surface of superquadric meshes. This representation retains the superquadric’s ability to parse a 3D scene into distinct parts. Meanwhile, spherical harmonics of Gaussians enable complex texture rendering via Gaussian splatting, addressing the texture learning limitations of prior work~\cite{liu2022robust,ramamonjisoa2022monteboxfinder, monnier2023dbw,yu2019partnet}. Sharing pose parameters between superquadrics and 2D Gaussians further improves the representation's efficiency. 


The parametric representation is controlled by both primitive and Gaussian parameters, which are optimized simultaneously. Given a 3D scene $\mathcal{S}$, the proposed method decomposes it into multiple hybrid blocks, each consisting of a superquadric with associated Gaussians. Each scene is denoted by the hybrid representation: $\mathcal{S} = \mathcal{B}_{1}\cup...\mathcal{B}_{i}...\cup\mathcal{B}_{M}$, where $\mathcal{B}_{i}$ is the $i$-th hybrid block, and $M$ is the total number of blocks. Blocks are defined using manually designed parameters that control pose, opacity, scale, shape, and texture. These parameters are optimized via differentiable rendering to parse the target scene.

\noindent \textbf{Shape and Scale Parameters}. For each hybrid block $\mathcal{B}_{i}$, its geometry is controlled by superquadric parameters~\cite{barr1981superquadrics}. Specifically, there are two shape parameters $\epsilon_{1}, \epsilon_{2}$ that define its shape, along with three scale parameters $s_1, s_2, s_3$, which scale the 3D axes. Analogous to icosphere vertex positioning, superquadric vertex coordinates are computed by:
\begin{equation}
\footnotesize
    \mathbf{v} = [s_1\cos^{\epsilon_1}(\theta)\cos^{\epsilon_2}(\varphi); 
                    s_2\sin^{\epsilon_1}(\theta); 
                    s_3\cos^{\epsilon_1}(\theta)\sin^{\epsilon_2}(\varphi)],
    \label{eq1}
\end{equation}
where $\theta$ and $\varphi$ represent the azimuth and elevation defined in the spherical coordinate. The shape and scale parameters govern block deformation to learn part-aware geometry.

\noindent \textbf{Pose Parameters}. 
The pose of the $i$-th hybrid block is defined by its rotation $\mathbf{R}_i$ and translation $\mathbf{t}_i$. The vertex position from Eq.~\ref{eq1} is transformed from the local coordinate to world space as: $\hat{\mathbf{v}}^{j}_i = \mathbf{R}_i \mathbf{v}^{j}_i + \mathbf{t}_i$, where $j$ indexes the vertices. Previous approaches, including SuGaR~\cite{guedon2023sugar} and GaMeS~\cite{waczynska2024games}, position Gaussians on reconstructed meshes with independent pose and shape parameters to enhance appearance modeling. 
In contrast,  our method employs Gaussians to construct a differentiable rendering that bridges images and superquadrics. To achieve this, we couple the parameters of Gaussians with superquadrics.   

For a posed superquadric, Gaussian centers are uniformly sampled on triangular faces, with their poses determined by face vertices. Following GaMeS~\cite{waczynska2024games}, each Gaussian's rotation matrix $\mathrm{R}_v = [r_1, r_2, r_3]$ and scaling $\mathrm{S}_v$ are computed from vertex positions. Given a triangular face $V$ with vertices $v_1, v_2, v_3 \in \mathbb{R}^3$, orthonormal vectors are constructed such that $r_1$ aligns with the face normal, $r_2$ points from the centroid $m = \text{mean}(v_1,v_2,v_3)$ to $v_1$. $r_3$ is obtained by orthonormalizing the vector from the center to the second vertex concerning the existing $r_1$ and $r_2$ :

\begin{equation}
r_3 = \frac{{ort}(v_2-m; r_1,r_2)}{|| {ort}(v_2-m; r_1,r_2 ||},
\end{equation}
where ${ort}$ represents the orthonormalization in Gram-Schmidt~\cite{bjorck1994numerics}. For the scaling of 2D Gaussians, we use:
\begin{equation}
    \mathrm{S}_v = \text{diag}(s_v^2,s_v^3),
\end{equation}
where $s_v^2= c||m-v_1||$,  $s_v^3 = c<v_2, r_3>$, and $c$ is a size control hyperparameter. We place a fixed number of Gaussians on each triangular face.  These designs eliminate the need to learn the geometry of Gaussians, thereby enhancing the efficiency of the representation.


\noindent \textbf{Opacity Parameters}. We define the total number of hybrid blocks as $M$. However, a typical scene does not contain exactly $M$ blocks. Therefore, only the meaningful blocks are retained. To achieve this, we introduce a learnable parameter $\tau_i$ to represent each block's opacity. During the optimization, only blocks with $\tau_i$ greater than a certain threshold are retained. Note that Gaussians within the same block share the same $\tau$ for opacity in rasterization.

\noindent \textbf{Texture Parameters}. The texture is modeled using  2D Gaussians positioned on the surface of the superquadrics. Spherical harmonics of each Gaussian control the texture and are optimized for rendering view-dependent images~\cite{kerbl20233d}.

\subsection{Block-level Decomposition}\label{sec:block-opt}
This section describes the optimization of the hybrid representation. We observed that minimizing the rendering loss across multi-view images alone led to instability in positioning hybrid blocks. Therefore, several regularization terms are introduced to optimize the composition for maximal image formation consistency.

\subsubsection{Optimization}
In addition to the image rendering loss, we employ regularizers consisting of coverage, overlap, parsimony, and opacity entropy. The coverage loss encourages hybrid blocks to cover only meaningful regions; the overlap loss prevents blocks from overlapping; the parsimony loss regularizes the number of existing blocks; and the opacity entropy encourages block opacities close to binary.

\noindent \textbf{Rendering Loss}. The rendering loss is based on 3DGS~\cite{kerbl20233d}, where it combines an $L_1$ term with a D-SSIM term:


\begin{equation}
    \cL_{\text{ren}} = (1-\lambda)\ L_1+ \lambda  L_{\text{D-SSIM}}.
\end{equation}

\noindent \textbf{Coverage Loss}. 
The coverage loss ensures that the block set $\{\mathcal{B}_{1}\dots \mathcal{B}_{M}\}$ covers the object while preventing it from extending beyond its boundaries. To determine the 3D occupancy field based on the blocks, we first define the approximate signed distance of a 3D point $x$ to the $i$-th block, $D_{i}(x) = \impsq_i(x) -1 $. Here, $\impsq_i$ denotes the superquadric inside-outside function~\cite{barr1981superquadrics} associated with the block $i$. Consequently, $D_{i}(x) \leq 0$ if the point $x$ lies inside the $i$-th block, and $D_{i}(x) > 0$ if $x$ is outside the block. Inspired by NeRF~\cite{mildenhall2020nerf}, we sample a ray set $\mathcal{R}$ using ray-casting based on camera poses. Given the object mask, we associate each ray $r \in \mathcal{R}$ with a binary label: $l_r = 1$ if the ray $r$ is inside mask, otherwise $l_r = 0$. The coverage loss is defined as:
\begin{equation}
	\cL_{\text{cov}}(\cR) = 
		\sum_{r \in \cR} l_r L_{\text{cross}}(r) + (1 - l_r) L_{\text{non-cross}}(r).
\label{eq:coverage_loss}
\end{equation}
Here, $L_{\text{cross}}$ encourages rays to intersect with blocks, while $L_{\text{non-cross}}$ discourages rays from intersecting with blocks:

\begin{equation}
	L_{\text{cross}}(r) = 
		 \text{ReLU}\Big(
		 \min_{i \in \mathcal{M}} \min_{x_j^r \in \cX_r} D_i(x_j^r)\Big),
\label{eq:cross}
\end{equation}
\begin{equation}
	L_{\text{non-cross}}(r) = 
		 \text{ReLU}\Big(
		 \max_{i \in \mathcal{M}} \max_{x_j^r \in \cX_r} -D_i(x_j^r)\Big),
\label{eq:non-cross}
\end{equation}
where $x_j^r$ means the $j$-th sampled point along the ray $r$ and $\mathcal{M}=\{ 1, \dots M \}$. Intuitively, this implies that at least one sampled points $\cX_r$ along the ray $r$ inside the mask that lies within a certain block, while all points along the ray outside the mask should not belong to any block.

\noindent \textbf{Overlap Loss}.
We introduce a regularization term to prevent overlap between individual blocks. Given the difficulty of directly calculating block overlap, we adopt a Monte Carlo method similar to ~\cite{monnier2023dbw}. Specifically, we sample multiple 3D points in space and penalize those that lie inside more than $k$ blocks. Based on the superquadric inside-outside function, a soft occupancy function is defined as:
\begin{equation}
\mathcal{O}_i^x = \tau_i \text{ sigmoid}(\frac{ - D_i(x)}{\gamma}),
\label{eq:soft_occ}
\end{equation}
where $\gamma$ is a temperature parameter.   
Thus, for a set of $N$ sampled 3D points $\pointset$, the overlap loss is expressed as:
\begin{equation}
  \cL_{\text{over}} = \frac{1}{N} \sum_{x \in \pointset} \text{ReLU} \left(\sum_{i\in \mathcal{M}}\mathcal{O}_i^x - k \right).
  \label{eq:overlap}
\end{equation}

\noindent \textbf{Parsimony Loss}. To promote the use of the minimal number of blocks and achieve parsimony in decomposition, we introduce a regularization term that penalizes block opacity ($\tau$). This loss is defined as:
\begin{equation}
  \cL_{\text{par}} = \frac{1}{M} \sum_{i \in \mathcal{M}} \sqrt{\tau_i}.
  \label{eq:parloss}
\end{equation}

\noindent \textbf{Opacity Entropy Loss}. During optimization, the opacity of the block inside the object region tends to approach 1, while the opacity of the block outside the object region tends to approach 0. To facilitate this, we associate the opacities of blocks with the labeled rays and apply a cross-entropy loss 
 $L_{ce}$ between the block opacity and the mask labels, defined as follows:
\begin{equation}
    \cL_{\text{opa}}(\cR) = 
    \frac{1}{|\cR|} \sum_{r \in \cR} L_{ce}\left(\max_{i \in \mathcal{M}} \tau_i(x^r), l_r\right)~. 
  \label{eq:opaloss}
\end{equation}
Here, only points  $x^r$ inside the blocks are sampled, 
\textit{e.g.}, $ \left\{ x^r \in \cX_r \mid D_i(x^r) \leq 0 \right\}$.

The total loss is the weighted sum of the loss terms described above:
\begin{equation}
    \cL = \cL_{\text{ren}} + \lambda_{\text{cov}}\cL_{\text{cov}}+\lambda_{\text{over}}\cL_{\text{over}}+\lambda_{\text{par}}\cL_{\text{par}}+\lambda_{\text{opa}}\cL_{\text{opa}}.
\end{equation}

\subsubsection{Adaptive Number of Blocks.} 
Given that the number of parts may vary across different scenes, we allow the number of blocks to adjust during optimization adaptively.  Specifically, when the opacity of an existing block falls below a threshold $t$, this block is removed immediately. Furthermore, we explore a block-adding mechanism to dynamically incorporate new components, ensuring comprehensive coverage of target objects. Specifically, DBSCAN~\cite{schubert2017dbscan} is employed to cluster initial point clouds that are not covered by any blocks, where the point clouds are either randomly initialized or derived as by-products from COLMAP~\cite{schoenberger2016sfm}. For each point cloud cluster, a new block is introduced at its center, with the remaining parameters of the block initialized randomly.

The final block-level reconstruction generates a textured 3D scene represented by multiple superquadrics, with surface details rendered through Gaussians, as illustrated in Fig.~\ref{fig:pip}. 


\subsection{Point-level Decomposition} \label{sec:point-opt}
The primitive-based hybrid representation effectively decomposes the shape into parts but exhibits suboptimally for complex objects. To address this limitation, we perform a refinement stage that enhances geometric fitting. In this stage, the constraint between Gaussians and superquadrics is decoupled, enabling independent Gaussian optimization.  Furthermore, to prevent Gaussian components from one block passing through adjacent blocks and disrupting the continuity and plausibility of the part decomposition, we impose a new constraint to minimize the negative signed distance of each Gaussian entering other blocks:
\begin{equation}
  \cL_{\text{enter}} = \frac{1}{N} \sum_{x \in \pointset} \sum_{m \in \mathcal{M} \setminus \{\delta\}} \text{ReLU}(-D_m(x)),
  \label{eq:refine}
\end{equation}
where $\pointset$ is the sampled Gaussian set, and $\delta$ represents the identifier of the block to which the Gaussian $x$ belongs. As seen in Fig.~\ref{fig:pip}, the reconstruction becomes more accurate and aligns better with the target shape.
\begin{figure*}[tbh]
  \centering
  \includegraphics[width=1.0\textwidth]{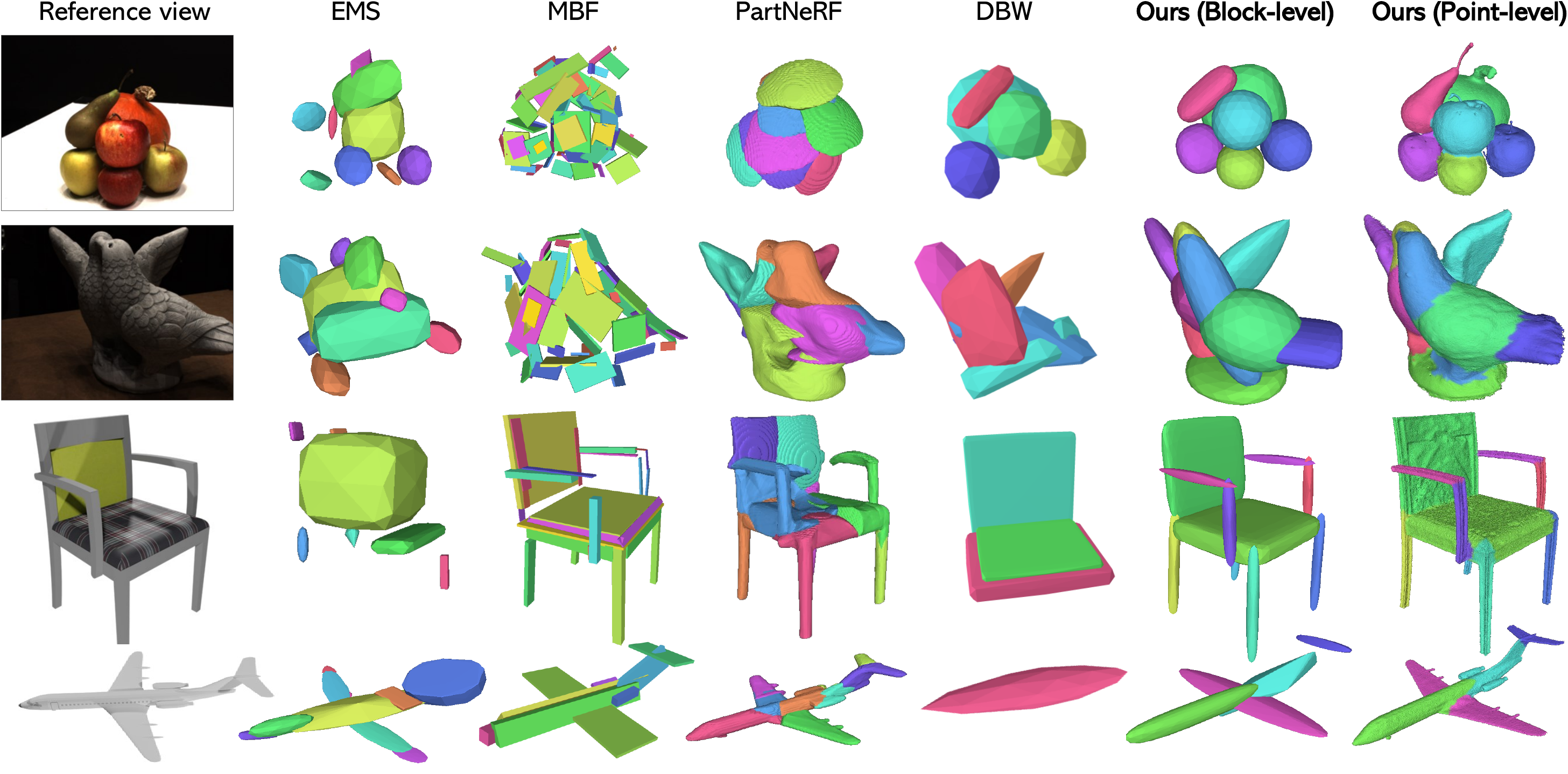}
  \caption{\textbf{Qualitative comparison on DTU~\cite{jensen2014large} and ShapeNet~\cite{jensen2014large} .} The first two rows are DTU examples, and the last two are ShapeNet examples, respectively.  Our method is the only one that provides reasonable 3D part decomposition while capturing detailed geometry.}
  \label{fig:dtu-shapenet-part}
\end{figure*}


\begin{figure}[!h]
  \centering
  \includegraphics[width=\linewidth]{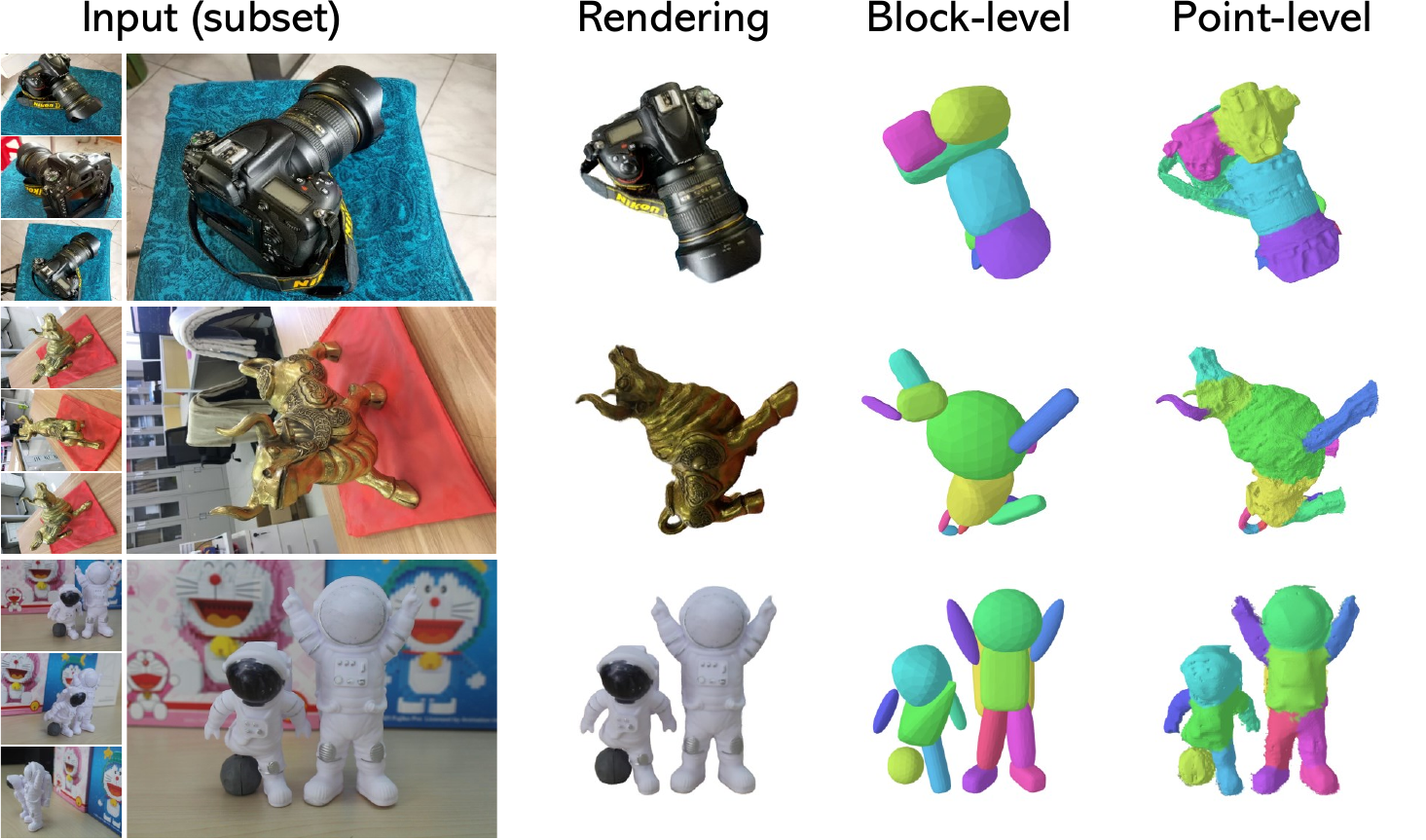}
  \caption{\textbf{Qualitative results on BlendedMVS~\cite{yao2020blendedmvs} and self-captured data.}  We present RGB  renderings and decomposed parts from novel views. The top examples are from the BlendedMVS dataset, and the last example is from our captured scenes.}
  \label{fig:mvs}
\end{figure}

\begin{figure}
  \centering
  \includegraphics[width=\linewidth]{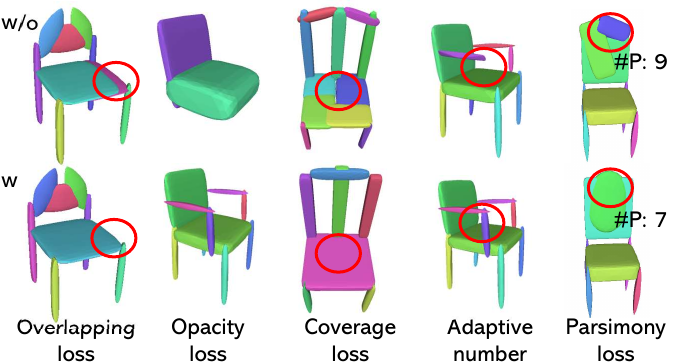}
  \caption{\textbf{Ablation studies on key strategies.} The block-level visual comparisons illustrate the impact of adopting our proposed strategy. The first row shows results without the strategy, and the second with the strategy implemented.}
   \vspace{-8px}
  \label{fig:ablation}
\end{figure}

\begin{table*}[t]
  \centering

  \vspace{.05in}
  {
\resizebox{\linewidth}{!}{\begin{tabular}{lccccccccccccccccccc}
  \toprule
  
  & & & \multicolumn{15}{c}{Chamfer distance per scene} & Mean & Mean \\
  \cmidrule(lr){4-18}

  Method & Input  & Renderable &   S24 & S37 &S40 & S55& S63 & S65 & S69 & S83 & S97 &
   S105 & S106 & S110  & S114  & S118  & S122 & CD & \#P \\
  \midrule
  EMS~\cite{liu2022robust} & 3D GT & \ding{55}&
  6.32 & 3.54 & 2.99 & 4.30 & 4.16 & 4.01 & 3.75 & 3.24 & 4.97 & 4.34& 4.16& 7.62& 7.58 &4.46&4.03& 4.65& 
   7.7 \\

  MBF~\cite{ramamonjisoa2022monteboxfinder} & 3D GT &\ding{55}  &
  3.12 & 2.66 & 3.84 & 2.54 & 1.59 & 2.11 & 2.19 & 2.01 & 2.32 & 2.45& 2.17& 2.12& 3.83 &2.02&2.55& 2.50& 
   \underline{34.1}   \\

  \midrule
  EMS~\cite{liu2022robust} + Neus~\cite{wang2021neus} & Image & \ding{55} &
  5.99 & 5.56 & 4.43 & 4.32 & 5.42 & 6.14 & 3.75 & 3.96 & 4.63 &  4.34& 5.88& 5.11 &4.29&4.83&3.53& 4.97& 
   8.87  \\

  MBF~\cite{ramamonjisoa2022monteboxfinder} + Neus~\cite{wang2021neus} & Image & \ding{55} &
  2.69 & 3.37 & 3.22 & 2.69 & 3.63& 2.60& 2.59 & 3.13 & 2.85 & 2.51& 2.45& 3.72& 2.24 &2.49&2.52& 2.85& 
   \underline{46.7}   \\

   PartNeRF~\cite{tertikas2023partnerf} & Image  &\checkmark &
  9.38 & 10.46 &9.08  &8.63  & 6.04& 7.25& 7.22& 9.15& 8.72 & 10.01 &6.72 &9.85 & 7.85& 8.68 & 9.21 &8.54 & 
  8.0   \\

  DBW~\cite{monnier2023dbw} & Image &\checkmark & 
  5.41 & 8.35 & 1.57 & 3.08 & 3.40 & 4.15 & 7.46 & 3.94 & 6.63 & 4.85& 4.38& 4.65&6.29& 4.34 &3.04&4.76 & 
  4.8  \\

\textbf{Ours (Block-level)}& Image &\checkmark &
  5.68 &  4.91 &1.85 & 2.61 & 3.75 & 4.66 & 3.75 & 7.57 & 4.27 & 4.38 & 3.49& 4.48& 3.61 &4.21 &3.70 & 4.19 &    5.9  \\

   \textbf{Ours (Point-level)} & Image  &\checkmark &
 \textbf{  0.70}   &\textbf{ 1.17} &\textbf{ 0.55 }& \textbf{0.65} & \textbf{1.06} & \textbf{1.23} &\textbf{1.10} &\textbf{1.36} & \textbf{1.37} & \textbf{0.78}& \textbf{0.92}&\textbf{1.41}& \textbf{0.69}&\textbf{1.05}&\textbf{0.71}& \textbf{0.98}&
  5.9  \\
  \bottomrule
  \end{tabular}
}
  \caption{\looseness=-1 \textbf{Quantitative comparison on DTU~\cite{jensen2014large}.} The Chamfer distance between the  3D reconstruction and the ground-truth is reported in 15 scenes.  The best results are bolded, and the average numbers of primitives found (\#P) that are greater than 10 are underlined.}
 \vspace{-10px}
  \label{tab:dtu_1}
}
\end{table*}

\begin{table}

\centering

 \vspace{.05in}

\resizebox{0.47\textwidth}{!}{\begin{tabular}{lcccccc}
  \toprule
Method & Part-aware & CD~$\downarrow$ & PSNR~$\uparrow$ & SSIM ~$\uparrow$ & LIPPS $\downarrow$  &Time~$\downarrow$ \\
\hline
Neuralangelo~\cite{li2023neuralangelo}& \ding{55}&{0.61} & {33.84}& -&- &   $\textgreater$ 10~h \\
2DGS~\cite{Huang2DGS2024}& \ding{55}&{0.81} & {34.07}& {0.99}&{0.019} &   $\sim$ 10~m \\

 \hline
 
PartNeRF~\cite{tertikas2023partnerf}&\checkmark & 9.59 & 17.97 & 0.77 & 0.246  & $\sim$ 8~h \\
DBW~\cite{monnier2023dbw}&\checkmark  & 4.73 & 16.44 & 0.75 & 0.201 &  $\sim$ 2~h \\
\textbf{Ours (Block-level)}&\checkmark& \second{4.19} &\second{19.84}&\second{0.82}&\second{0.189}&   $\sim$ 30~m \\

\textbf{Ours (Point-level) }&\checkmark& \textbf{0.98} & \textbf{35.04} & \textbf{0.99} & \textbf{0.015} &  $\sim$ 40 m \\
  \bottomrule
\end{tabular}}

\caption{\textbf{Quantitative results on DTU}~\cite{jensen2014large}. 
Our method outperforms all part-aware approaches in image synthesis quality, reconstruction accuracy, and efficiency. Neuralangelo's results are from the original paper, with all times measured on an RTX 3090 GPU. }
  \vspace{-10px}
\label{tab:dtu_2}
\end{table}

\begin{table}
\small
  \centering
  \addtolength{\tabcolsep}{-2.9pt}

  \vspace{.05in}
  \resizebox{0.48\textwidth}{!}{%
    \begin{tabular}{lccccc|cccccc}
        \toprule
         &  &
        \multicolumn{4}{c|}{ Chamfer Distance ~$\downarrow$} & \multicolumn{4}{c}{Primitives (\#P )} & Mean & Mean \\ \cmidrule{3-6} \cmidrule{7-10} 
       Method & Input & Airplane & Table & Chair & Gun & Airplane & Table & Chair & Gun & CD & \#P \\
        \midrule
        EMS~\cite{liu2022robust} & 3D GT & 3.40 &6.92  &19.0  &2.02    & 9.4 & 7.88 &  10.3 & 8.4 &7.84 &  9.0 \\

          MBF~\cite{ramamonjisoa2022monteboxfinder} & 3D GT & 2.83 & 2.18 & \second{1.59}&   2.32 &  10.85 & 13.9 &  13.4  & 14.3  &2.21 & 13.1  \\
        \midrule
        PartNeRF~\cite{tertikas2023partnerf} & Image & \second{2.29} & 2.77 & 2.30 &  2.46 & 8.0 & 8.0 &8.0 & 8.0 & 2.46  & 8.0  \\
        DBW~\cite{monnier2023dbw} & Image & 3.61 & 7.33 &  6.19& 2.09 &  2.7 & 5.2& 3.6 &3.3  & 4.81 &  3.7 \\
        \textbf{Ours (Block-level)} & Image & 2.47 & \second{2.15}  & 2.32 & \second{1.78}  &3.9  & 6.6 &7.6  &5.0  & \second{2.18}  &  5.8 \\
       \textbf{Ours (Point-level)} & Image & \first{1.29} & \first{1.72} & \first{0.94} & \first{1.07}  & 3.9 & 6.6 & 7.6 &5.0  & \first{1.25}  &  5.8 \\
        \bottomrule
    \end{tabular}
  }
    \caption{\looseness=-1 \textbf{Quantitative comparison on ShapeNet~\cite{chang2015shapenet}.} We report Chamfer distance and the number of parts. The best results are bolded, and the second-best results are underlined. 
    }   
     \vspace{-4px}
  \label{tab:shapenet}
\end{table}

\begin{table*}

 \vspace{.05in}
{
 \resizebox{1.0\textwidth}{!}{%
    \begin{tabular}{lccccc|cccc|cccc|cccc}
        \toprule
         &  &
        \multicolumn{4}{c|}{Chamfer Distance ~$\downarrow$ } & \multicolumn{4}{c|}{PSNR ~$\uparrow$} & \multicolumn{4}{c|}{SSIM ~$\uparrow$}& \multicolumn{4}{c}{LIPPS ~$\downarrow$}  \\ \cmidrule{3-6} \cmidrule{7-10} \cmidrule{11-14}  \cmidrule{15-18} 
       Method & Part-aware & Airplane & Table & Chair & Gun & Airplane & Table & Chair & Gun & Airplane & Table & Chair & Gun & Airplane & Table & Chair & Gun \\
       \hline
2DGS~\cite{Huang2DGS2024}& \ding{55}& 1.47 & 2.37 & 0.50 & 1.03 & 40.89&39.80&39.05&41.74& 0.994&0.990&0.990&0.994& 0.009&0.027&0.017&0.009 \\
 \hline
PartNeRF~\cite{tertikas2023partnerf}&\checkmark & \second{2.29} & 2.77 & \second{2.30} &  2.46 & 19.63 & 20.66 & 19.08 & 21.97 &  0.898 & 0.855 & 0.875 & 0.916 & 0.086 & 0.161 & 0.136 & 0.083\\
DBW~\cite{monnier2023dbw}&\checkmark &3.61 & 7.33 & 6.19& 2.09&26.11&23.84&20.25&{28.72} &{0.950}&{0.915}&{0.892}&0.960&{0.074}&{0.136}&0.132&\second{0.042} \\

\textbf{Ours (Block-level)}&\checkmark & 2.47 &\second{2.15}  & 2.32 & \second{1.78} &\second{27.94} &\second{27.98}&\second{24.92}&\second{29.95}&\second{0.959}&\second{0.925}&\second{0.906}&\second{0.963}& \second{0.072}&\second{0.129}&\second{0.122}&{0.047} \\
\textbf{Ours (Point-level)}&\checkmark& \first{1.29} & \first{1.72} & \first{0.94} & \first{1.07} &\first{41.18}&\first{36.80} &\first{36.07}&\first{39.51} &\first{0.992}&\first{0.973}&\first{0.977}&\first{0.989}&\first{0.014}&\first{0.070}&\first{0.038}&\first{0.021} \\
  \bottomrule
    \end{tabular}
}
\centering
\caption{\textbf{Quantitative results on ShapeNet}~\cite{chang2015shapenet}. We report the Chamfer distance and novel view synthesis results across four categories. }
\label{tab:shapenet_2}
  \vspace{-4px}

}
\end{table*}


\section{Experiments}
We comprehensively evaluate our approach across four key aspects: 3D reconstruction quality, view synthesis performance, shape parsimony, and computational efficiency, comparing it with state-of-the-art methods in the field.
\subsection{Datasets}
Extensive experiments are conducted on two widely-used public datasets: DTU \cite{jensen2014large} and ShapeNet \cite{chang2015shapenet}. Specifically, 15 standard scenes from the DTU dataset that are widely adopted for reconstruction quality assessment are used for evaluation. Each scene contains either 49 or 64 images. Additionally, we construct a ShapeNet subset consisting of four categories: \textit{Chair}, \textit{Table}, \textit{Gun}, and \textit{Airplane}, with 15 objects selected per category.  For each object, 50 rendered images are generated for training.  We further validate real-world applicability on both the BlendedMVS dataset~\cite{yao2020blendedmvs} and self-captured scenes, demonstrating robust performance across synthetic and natural environments.

\subsection{Implementation Details} 
We build upon the 2DGS \cite{Huang2DGS2024} architecture and add a non-learnable part attribute to each Gaussian component to enforce the constraint in Eq.~\ref{eq:refine}. In the ShapeNet benchmark, we observe a common degradation in mesh reconstruction as shown in Fig.~\ref{fig:ablation_sr}. To address this,  we propose Gaussian scale regularization, as validated in Sec.~\ref{sec:ablation}.   We set the initial number of primitives $M$ to 8.  The temperature parameter $\gamma$ in Eq.~\ref{eq:soft_occ} is 0.005 and the overlapping number $k$ in Eq.~\ref{eq:overlap} is 1.95. We train the hybrid representation for 30k iterations, followed by refinement optimization for another 30k iterations. During block-level optimization, the block-adding operation is carried out at the 5k-th and 10k-th iterations. The loss weights $\lambda_\text{cov}$, $\lambda_\text{over}$, $\lambda_\text{par}$, and $\lambda_\text{opa}$ are set to 10, 1, 0.002, and 0.01, respectively. For more details about datasets, metrics, and baselines, please refer to \textit{Supplementary}.

\subsection{Evaluations}
\subsubsection{DTU and Shapenet Benchmark} \label{sec:dtu}

\textit{Geometry Reconstruction.} In Tab.~\ref{tab:dtu_1} and Tab.~\ref{tab:shapenet}, we compare our geometry reconstruction to SOTA shape decomposition methods on Chamfer distance and training time using DTU and Shapenet dataset. The Chamfer distance metrics reported for ShapeNet are scaled by a factor of 100 for readability.  Our method consistently outperforms in all scenes compared to prior works.  As shown in Fig.~\ref{fig:dtu-shapenet-part}, our approach consistently produces interpretable 3D decompositions with further refinement achieving more detailed geometry (the last two columns). MBF~\cite{ramamonjisoa2022monteboxfinder} achieves a low CD error but at the cost of using significantly more primitives, leading to over-decomposition and the inclusion of meaningless parts. Moreover, as shown in Tab.~\ref{tab:dtu_2} and Tab.~\ref{tab:shapenet_2},  our approach achieves competitive results with the advanced Gaussian-based 2DGS~\cite{Huang2DGS2024} and surface reconstruction Neuralangelo~\cite{li2023neuralangelo}, which are limited to producing unstructured meshes.  Notably, our model demonstrates excellent efficiency, with a reconstruction speed approximately 10X faster than PartNeRF~\cite{tertikas2023partnerf} and over 3X faster than DBW~\cite{monnier2023dbw}. 

\textit{Appearance Reconstruction.} In addition to part decomposition, our method enables high-fidelity image synthesis, attributed to integrating Gaussian splatting within hybrid representations.
EMS~\cite{liu2022robust} and MBF~\cite{ramamonjisoa2022monteboxfinder} operate directly on point clouds and consequently lack image rendering capabilities, while PartNeRF~\cite{tertikas2023partnerf} and DBW~\cite{monnier2023dbw} yield low-quality view synthesis. In contrast, our method achieves high-quality appearance rendering results, as demonstrated in Tab.~\ref{tab:dtu_2} and Tab.~\ref{tab:shapenet_2}.

\begin{figure}[t]
  \centering
  \includegraphics[width=\linewidth]{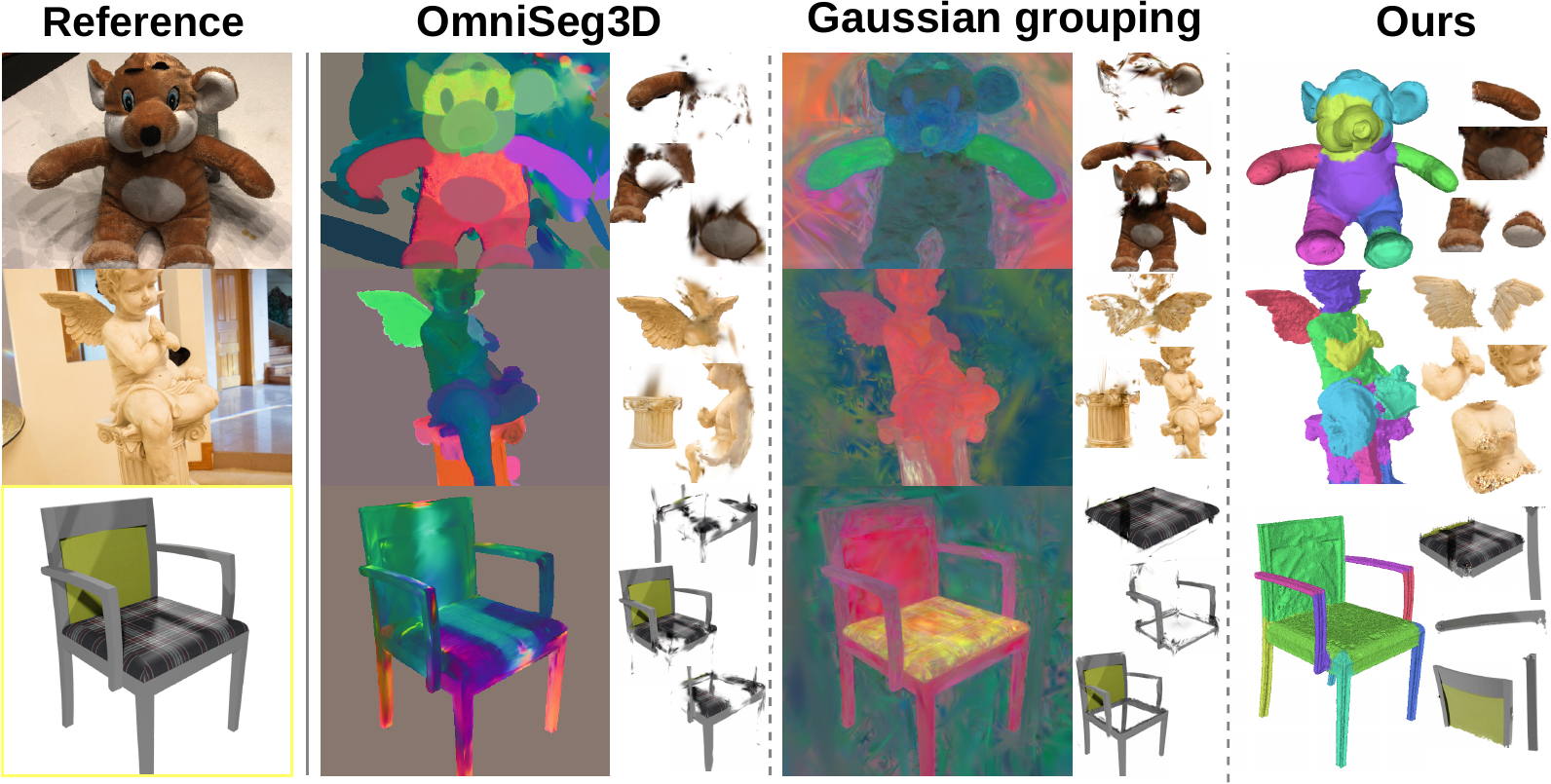}
  \caption{\textbf{Qualitative comparison with SAM-based methods}. Visual results demonstrate that our method produces more structurally coherent decompositions, whereas SAM-based approaches frequently exhibit spatial discontinuities.}
  \label{fig:sam}
\end{figure}

  







\begin{table}
  \addtolength{\tabcolsep}{-1pt}
  
  \centering

\resizebox{\linewidth}{!}{
\begin{tabular}{lCCCCCCCCCC}
  \toprule
  
  & \multicolumn{4}{c}{Block-level} & \multicolumn{4}{c}{Point-level} &  \\
  \cmidrule(lr){2-5} \cmidrule(lr){6-9}
  Method & \text{CD}\downarrow & \text{PSNR}\uparrow & \text{SSIM}\uparrow & \text{LPIPS}\downarrow  & \text{CD}\downarrow & \text{PSNR}\uparrow & \text{SSIM}\uparrow & \text{LPIPS}\downarrow & \text{\#P} \\
  \midrule

  \textbf{Complete model }  & 2.32 & 24.92 & 0.906 & 0.122  & \underline{0.94} & \underline{36.07} & 0.977 & 0.038  & 6.6 \\

  w/o $L_{\text{over}}$  & \underline{2.11} & \underline{25.17} & \underline{0.914} & \underline{0.118}  & 1.34 & 36.04 & \underline{0.977} & \underline{0.037} & 7.6  \\

  w/o $L_{\text{opa}}$  & 5.72 & 21.12 & 0.853 & 0.174  & 1.06 & 35.60 & 0.975 & 0.040  & 3.8 \\

  w/o $L_{\text{cov}}$  & 3.21 & 22.96 & 0.896 & 0.134  & 1.34 & 36.00 & 0.976 & 0.038  & 8.53 \\

  w/o Adaptive  & 3.16 & 22.74 & 0.880 & 0.141  & 1.05 & 35.74 & 0.974 & 0.040  & 6.7 \\

   w/o $L_{\text{par}}$  & \textbf{1.91} & \textbf{25.48} & \textbf{0.918} & \textbf{0.115}  & \textbf{0.93} & \textbf{36.18} & \textbf{0.977} & \textbf{0.037}  & 10.1 \\

  \bottomrule
  \end{tabular}
  }

  \caption{\textbf{Ablation studies on the ShapeNet~\cite{chang2015shapenet}.} We report Chamfer Distance, rendering metrics, and the number of primitives (\#P). }  
    \label{tab:ablation_detailed}
\end{table}

\subsubsection{Real-life Data}
To further demonstrate the applicability of our method for learning shape decomposition, we test our model on the real-life images from the BlendedMVS dataset~\cite{yao2020blendedmvs} and a self-captured dataset.  As shown in Fig.~\ref{fig:mvs}, our approach can robustly produce both realistic appearances and reasonable 3D decompositions across a variety of data types. More results are provided in the supplementary material.


\begin{figure}
  \centering
  \includegraphics[width=\linewidth]{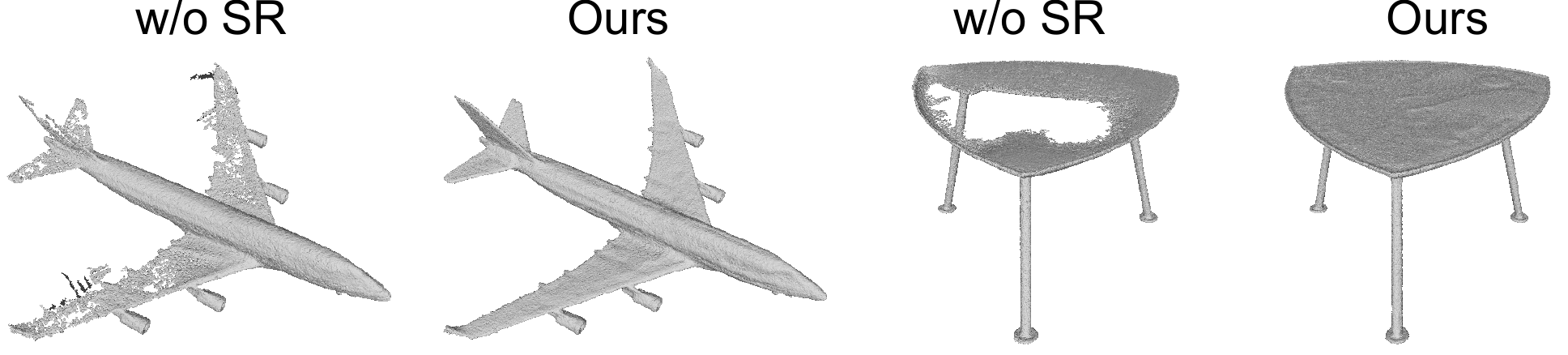}
 \caption{\textbf{Impact of Gaussian scale regularization (SR).} 
 The degraded mesh produced by 2DGS~\cite{Huang2DGS2024} is effectively improved.
 }
  \label{fig:ablation_sr}
   \vspace{-6px}
\end{figure}

\begin{figure}
  \centering
  \includegraphics[width=\linewidth]{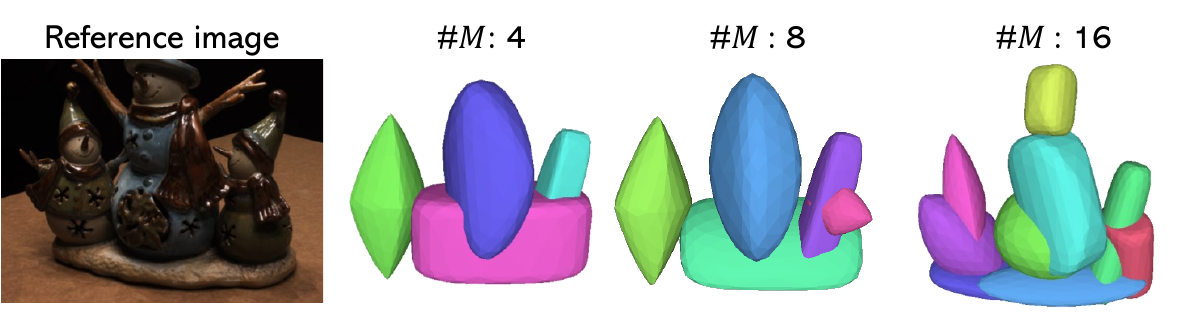}
 \caption{\textbf{Impact of the initial number of primitives ($M$).}  Increasing $M$ yields a finer-grained decomposition, while decreasing it produces a coarser decomposition.}
  \label{fig:granularity}
   \vspace{-6px}
\end{figure}

\begin{figure}
  \centering
  \includegraphics[width=1.0\linewidth]{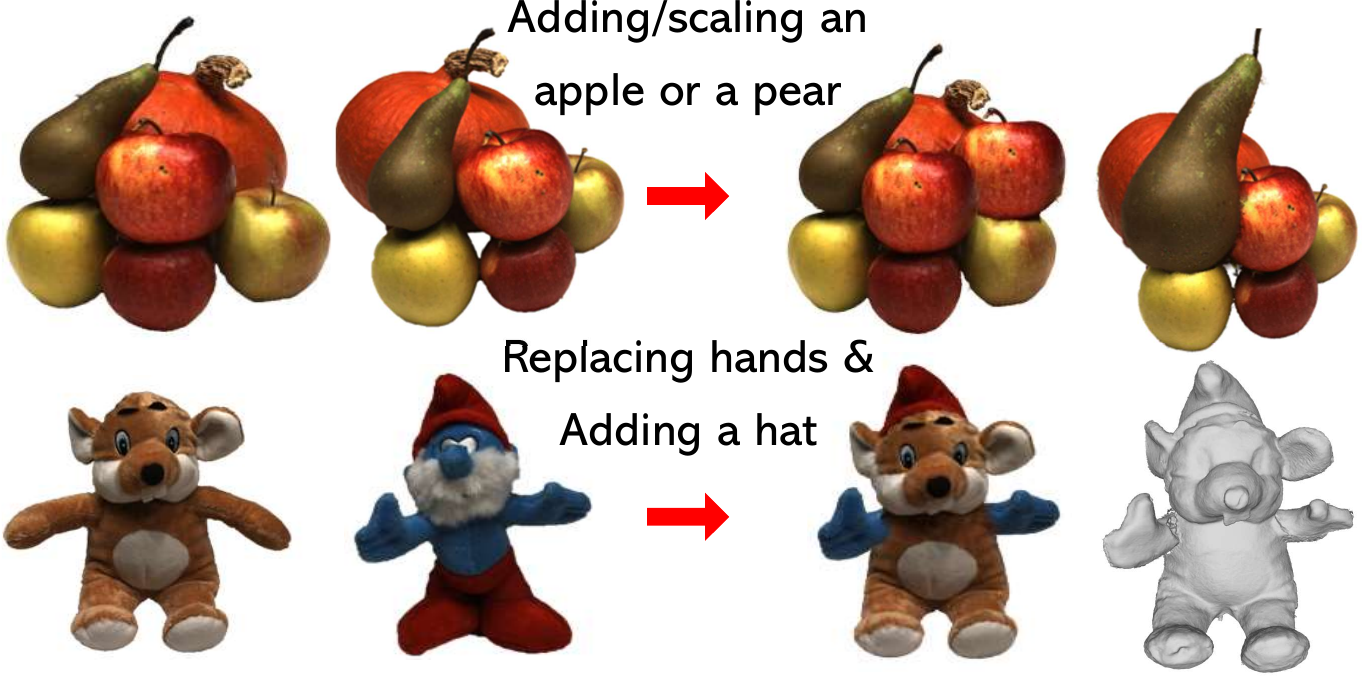}
  \caption{\textbf{Applications}. \textbf{Part-Aware Editing (top):} After optimization, we can easily edit the scene by adding, scaling, or moving specific parts. \textbf{ 3D Content Generation (bottom): } By combining parts of different objects, we can create new 3D content.}
  \label{fig:application}
   \vspace{-5px}
\end{figure}

\subsubsection{Compared to SAM-based Methods} We also conduct comparisons to SAM-based~\cite{kirillov2023segany} methods for this task, as shown in Fig.~\ref{fig:sam}. Despite impressive advances in 3D segmentation and editing~\cite{cen2023saga, garfield2024, gaussian_grouping, ying2023omniseg3d} by distilling 2D information, achieving semantic disentanglement but remains challenging due to indistinct textures and severe cross-view 2D inconsistencies.  Lifting 2D segmentation features to 3D segmentation will also lead to feature misalignment, leading to visual artifacts. In contrast, our method performs direct 3D space segmentation, yielding more reasonable and structurally coherent decompositions.

\subsection{Ablations}

\label{sec:ablation}

\begin{table}
  \small
  \addtolength{\tabcolsep}{-1pt}

  \centering

  \vspace{.05in}
\resizebox{\linewidth}{!}{
\begin{tabular}{lCCCCCCCCCC}
  \toprule
  
  & \multicolumn{4}{c}{Block-level} & \multicolumn{4}{c}{Point-level} &  \\
  \cmidrule(lr){2-5} \cmidrule(lr){6-9}
  Method & \text{CD}\downarrow & \text{PSNR}\uparrow & \text{SSIM}\uparrow & \text{LPIPS}\downarrow  & \text{CD}\downarrow & \text{PSNR}\uparrow & \text{SSIM}\uparrow & \text{LPIPS}\downarrow & \text{\#P} \\
  \midrule

   $\lambda_\text{par}$ = 0.1  & 6.46 & 16.10 & 0.721 & 0.255  & 1.08 & 34.30 & 0.987 & 0.018  & 2.7\\

$\lambda_\text{par}$ = 0.01 (*)   &  4.19 & 19.84 & 0.820 & 0.189  & 1.05 & 35.04 & 0.988 & 0.015 & 5.9 \\
  
 $\lambda_\text{par}$ = 0.001   &  4.01 & 20.18 & 0.835 & 0.180  & 1.03 & 35.04 & 0.988 & 0.015  & 7.1 \\

  \midrule
   $M$= 4  & 4.97 & 19.06 & 0.794 & 0.213  & 1.06 & 34.93 & 0.987 & 0.016 & 5.0 \\

 $M$= 8 (*)  &  4.19 & 19.84 & 0.820 & 0.189  & 1.05 & 35.04 & 0.988 & 0.015 & 5.9 \\
  
  $M$= 16  &  3.99 & 20.87 & 0.83 & 0.176  & 1.07 & 35.05& 0.988 & 0.015 & 8.3\\
    \bottomrule
  \end{tabular}
  }
    \caption{\textbf{
  Effect of parsimony weight ($\lambda_\text{par}$) and initial primitives count ($M$) on DTU~\cite{jensen2014large}.} We report Chamfer distance, rendering metrics, and primitives count (\#P). * denotes the default setting.}
  
    \label{tab:ablation_weight}
    \vspace{-15px}
\end{table} 

In this section, we first analyze the design choices by isolating each strategy to assess its impact. We report the averaged quantitative performance on 15 instances of the chair category in Tab.~\ref{tab:ablation_detailed} and provide visual comparisons in  Fig.~\ref{fig:ablation}.  Removing overlapping loss provides better reconstruction accuracy but more overlapping parts. The opacity loss drastically affects the number of primitives and doubles the CD, highlighting its role in controlling primitive presence and reconstruction quality.  Coverage loss is crucial in reconstruction accuracy, ensuring primitives align correctly with the target. The adaptive primitive strategy fills in missing parts and improves the reconstruction metric. Without parsimony loss, CD and rendering metrics are optimal but result in over-decomposition. Note that the full model did not yield the highest accuracy because the part-aware reconstruction trades off the accuracy and rationality of the part decomposition. The objective is to achieve high accuracy while preserving reasonable parts.

We also validate the effectiveness of Gaussian scale regularization in Fig.~\ref{fig:ablation_sr}.  2DGS~\cite{Huang2DGS2024} tends to use large-scale Gaussians in texture-less areas and create holes in the constructed mesh by TSDF integration~\cite{Zhou2018}. The presence of holes stems from the inability of large-scale Gaussians to maintain view-consistent depth.  Scale regularization effectively addresses this issue by suppressing large-scale Gaussians.

Lastly, in Tab.~\ref{tab:ablation_weight}, we analyze the influence of two key hyperparameters on decomposition granularity: the parsimony loss weight $\lambda_\text{par}$ and the initial primitive count $M$. Stronger parsimony regularization reduces the number of primitives, while weaker regularization increases it. As $M$ rises, both reconstruction and view synthesis performance slightly improve. This demonstrates that adjusting $M$ and $\lambda_\text{par}$ enables our method to effectively control the granularity of object or scene decomposition. Fig.~\ref{fig:granularity} visually illustrates this impact.

\subsection{Applications}
Fig.~\ref{fig:application} illustrates two applications of our method that the original 3DGS and NeRF-based approaches do not support. Firstly, after optimization, we obtain the part decomposition, facilitating easy editing of specific objects or scene components, e.g., adding, moving, removing, or scaling. Secondly, by combining parts of different objects, our method enables the creation of new high-quality 3D content. 

\section{Conclusions}

We introduce PartGS, a hybrid representation of superquadrics and 2D Gaussians, to learn 3D scenes in a part-aware representation. Compared to prior works, the proposed method retains geometry details, supporting high-quality image rendering.  It obtains state-of-the-art performance in comprehensive evaluations. One limitation is that it takes background-free scene images as inputs, leveraging segmentation tools such as SAM~\cite{kirillov2023segany}. 
In the future, we aim to explore how to model backgrounds and extend to larger and more complex scenes.

\section*{Acknowledgements}
This work is supported in part by the NSFC (62325211, 62132021, 62372457), the Major Program of Xiangjiang Laboratory (23XJ01009), Young Elite Scientists Sponsorship Program by CAST (2023QNRC001), the Natural Science Foundation of Hunan Province of China (2022RC1104).
{
    \small
    \bibliographystyle{ieeenat_fullname}
    \bibliography{main}

\begin{thebibliography}{72}
\providecommand{\natexlab}[1]{#1}
\providecommand{\url}[1]{\texttt{#1}}
\expandafter\ifx\csname urlstyle\endcsname\relax
  \providecommand{\doi}[1]{doi: #1}\else
  \providecommand{\doi}{doi: \begingroup \urlstyle{rm}\Url}\fi

\bibitem[Alaniz et~al.(2023)Alaniz, Mancini, and Akata]{Alaniz_2023_ICCV}
Stephan Alaniz, Massimiliano Mancini, and Zeynep Akata.
\newblock Iterative superquadric recomposition of 3d objects from multiple views.
\newblock In \emph{Proceedings of the IEEE/CVF International Conference on Computer Vision (ICCV)}, pages 18013--18023, 2023.

\bibitem[Barr(1981)]{barr1981superquadrics}
Alan~H. Barr.
\newblock Superquadrics and {{Angle-Preserving Transformations}}.
\newblock \emph{IEEE Computer Graphics and Applications}, 1981.

\bibitem[Binford(1971)]{binford1971visual}
Thomas Binford.
\newblock {Visual Perception by Computer}.
\newblock In \emph{IEEE Conference on Systems and Control}, 1971.

\bibitem[Bj{\"o}rck(1994)]{bjorck1994numerics}
{\AA}ke Bj{\"o}rck.
\newblock Numerics of gram-schmidt orthogonalization.
\newblock \emph{Linear Algebra and Its Applications}, 197:\penalty0 297--316, 1994.

\bibitem[Cen et~al.(2023)Cen, Fang, Yang, Xie, Zhang, Shen, and Tian]{cen2023saga}
Jiazhong Cen, Jiemin Fang, Chen Yang, Lingxi Xie, Xiaopeng Zhang, Wei Shen, and Qi Tian.
\newblock Segment any 3d gaussians.
\newblock \emph{arXiv preprint arXiv:2312.00860}, 2023.

\bibitem[Chang et~al.(2015)Chang, Funkhouser, Guibas, Hanrahan, Huang, Li, Savarese, Savva, Song, Su, Xiao, Yi, and Yu]{chang2015shapenet}
Angel~X. Chang, Thomas Funkhouser, Leonidas Guibas, Pat Hanrahan, Qixing Huang, Zimo Li, Silvio Savarese, Manolis Savva, Shuran Song, Hao Su, Jianxiong Xiao, Li Yi, and Fisher Yu.
\newblock {{ShapeNet}}: {{An Information-Rich 3D Model Repository}}.
\newblock \emph{arXiv:1512.03012 [cs.CV]}, 2015.

\bibitem[Chen et~al.(2023{\natexlab{a}})Chen, Li, and Lee]{chen2023neusg}
Hanlin Chen, Chen Li, and Gim~Hee Lee.
\newblock Neusg: Neural implicit surface reconstruction with 3d gaussian splatting guidance.
\newblock \emph{arXiv preprint arXiv:2312.00846}, 2023{\natexlab{a}}.

\bibitem[Chen et~al.(2020)Chen, Tagliasacchi, and Zhang]{Chen_Tagliasacchi_Zhang_2020}
Zhiqin Chen, Andrea Tagliasacchi, and Hao Zhang.
\newblock Bsp-net: Generating compact meshes via binary space partitioning.
\newblock In \emph{2020 IEEE/CVF Conference on Computer Vision and Pattern Recognition (CVPR)}, 2020.

\bibitem[Chen et~al.(2023{\natexlab{b}})Chen, Chen, Zhou, and Zhang]{chen2023dae}
Zhiqin Chen, Qimin Chen, Hang Zhou, and Hao Zhang.
\newblock Dae-net: Deforming auto-encoder for fine-grained shape co-segmentation.
\newblock \emph{arXiv preprint arXiv:2311.13125}, 2023{\natexlab{b}}.

\bibitem[Deng et~al.(2020)Deng, Genova, Yazdani, Bouaziz, Hinton, and Tagliasacchi]{Deng_Genova_Yazdani_Bouaziz_Hinton_Tagliasacchi_2020}
Boyang Deng, Kyle Genova, Soroosh Yazdani, Sofien Bouaziz, Geoffrey Hinton, and Andrea Tagliasacchi.
\newblock Cvxnet: Learnable convex decomposition.
\newblock In \emph{2020 IEEE/CVF Conference on Computer Vision and Pattern Recognition (CVPR)}, 2020.

\bibitem[Fischler and Bolles(1981)]{fischler1981random}
Martin~A. Fischler and Robert~C. Bolles.
\newblock Random {{Sample Consensus}}: {{A Paradigm}} for {{Model Fitting}} with {{Applications}} to {{Image Analysis}} and {{Automated Cartography}}.
\newblock \emph{Communications of the ACM}, 1981.

\bibitem[Gao et~al.(2024{\natexlab{a}})Gao, Shen, Zhang, Xiong, Peng, Gao, Wang, Xie, and Wang]{gao2024fdc}
Huachen Gao, Shihe Shen, Zhe Zhang, Kaiqiang Xiong, Rui Peng, Zhirui Gao, Qi Wang, Yugui Xie, and Ronggang Wang.
\newblock Fdc-nerf: learning pose-free neural radiance fields with flow-depth consistency.
\newblock In \emph{ICASSP 2024-2024 IEEE International Conference on Acoustics, Speech and Signal Processing (ICASSP)}, pages 3615--3619. IEEE, 2024{\natexlab{a}}.

\bibitem[Gao et~al.(2019)Gao, Yang, Wu, Yuan, Fu, Lai, and Zhang]{gao2019sdm}
Lin Gao, Jie Yang, Tong Wu, Yu-Jie Yuan, Hongbo Fu, Yu-Kun Lai, and Hao Zhang.
\newblock Sdm-net: Deep generative network for structured deformable mesh.
\newblock \emph{ACM Transactions on Graphics (TOG)}, 38\penalty0 (6):\penalty0 1--15, 2019.

\bibitem[Gao et~al.(2024{\natexlab{b}})Gao, Yang, Zhang, Sun, Yuan, Fu, and Lai]{gao2024mesh}
Lin Gao, Jie Yang, Bo-Tao Zhang, Jia-Mu Sun, Yu-Jie Yuan, Hongbo Fu, and Yu-Kun Lai.
\newblock Mesh-based gaussian splatting for real-time large-scale deformation.
\newblock \emph{arXiv preprint arXiv:2402.04796}, 2024{\natexlab{b}}.

\bibitem[Gao et~al.(2024{\natexlab{c}})Gao, Yi, Qin, Ye, Zhu, and Xu]{gao2024learning}
Zhirui Gao, Renjiao Yi, Zheng Qin, Yunfan Ye, Chenyang Zhu, and Kai Xu.
\newblock Learning accurate template matching with differentiable coarse-to-fine correspondence refinement.
\newblock \emph{Computational Visual Media}, 10\penalty0 (2):\penalty0 309--330, 2024{\natexlab{c}}.

\bibitem[Gao et~al.(2025{\natexlab{a}})Gao, Yi, Dai, Zhu, Chen, Zhu, and Xu]{gao2025curveawaregaussiansplatting3d}
Zhirui Gao, Renjiao Yi, Yaqiao Dai, Xuening Zhu, Wei Chen, Chenyang Zhu, and Kai Xu.
\newblock Curve-aware gaussian splatting for 3d parametric curve reconstruction, 2025{\natexlab{a}}.

\bibitem[Gao et~al.(2025{\natexlab{b}})Gao, Yi, Zhu, Zhuang, Chen, and Xu]{gao2025generic}
Zhirui Gao, Renjiao Yi, Chenyang Zhu, Ke Zhuang, Wei Chen, and Kai Xu.
\newblock Generic objects as pose probes for few-shot view synthesis.
\newblock \emph{IEEE Transactions on Circuits and Systems for Video Technology}, 2025{\natexlab{b}}.

\bibitem[Gu et~al.(2025)Gu, Cui, Li, Wei, Ge, Gu, Liu, Davis, and Ding]{Gu_2025_CVPR}
Zeqi Gu, Yin Cui, Zhaoshuo Li, Fangyin Wei, Yunhao Ge, Jinwei Gu, Ming-Yu Liu, Abe Davis, and Yifan Ding.
\newblock Artiscene: Language-driven artistic 3d scene generation through image intermediary.
\newblock In \emph{Proceedings of the Computer Vision and Pattern Recognition Conference (CVPR)}, pages 2891--2901, 2025.

\bibitem[Guan et~al.(2020)Guan, Liu, Liu, Yin, Hu, van Kaick, Zhang, Yumer, Carr, Mech, et~al.]{guan2020fame}
Yanran Guan, Han Liu, Kun Liu, Kangxue Yin, Ruizhen Hu, Oliver van Kaick, Yan Zhang, Ersin Yumer, Nathan Carr, Radomir Mech, et~al.
\newblock Fame: 3d shape generation via functionality-aware model evolution.
\newblock \emph{IEEE Transactions on Visualization and Computer Graphics}, 28\penalty0 (4):\penalty0 1758--1772, 2020.

\bibitem[Gu{\'e}don and Lepetit(2023)]{guedon2023sugar}
Antoine Gu{\'e}don and Vincent Lepetit.
\newblock Sugar: Surface-aligned gaussian splatting for efficient 3d mesh reconstruction and high-quality mesh rendering.
\newblock \emph{arXiv preprint arXiv:2311.12775}, 2023.

\bibitem[Huang et~al.(2024)Huang, Yu, Chen, Geiger, and Gao]{Huang2DGS2024}
Binbin Huang, Zehao Yu, Anpei Chen, Andreas Geiger, and Shenghua Gao.
\newblock 2d gaussian splatting for geometrically accurate radiance fields.
\newblock \emph{SIGGRAPH}, 2024.

\bibitem[Hui et~al.(2022)Hui, Li, Hu, and Fu]{hui2022neural}
Ka-Hei Hui, Ruihui Li, Jingyu Hu, and Chi-Wing Fu.
\newblock Neural template: Topology-aware reconstruction and disentangled generation of 3d meshes.
\newblock In \emph{Proceedings of the IEEE/CVF conference on computer vision and pattern recognition}, pages 18572--18582, 2022.

\bibitem[Jensen et~al.(2014)Jensen, Dahl, Vogiatzis, Tola, and Aanaes]{jensen2014large}
Rasmus Jensen, Anders Dahl, George Vogiatzis, Engil Tola, and Henrik Aanaes.
\newblock Large {{Scale Multi-view Stereopsis Evaluation}}.
\newblock In \emph{{{CVPR}}}, 2014.

\bibitem[Jiang et~al.(2024)Jiang, Zhao, Rahmani, Soh, Liu, and Zhao]{jiang2024gaussianblock}
Shuyi Jiang, Qihao Zhao, Hossein Rahmani, De~Wen Soh, Jun Liu, and Na Zhao.
\newblock Gaussianblock: Building part-aware compositional and editable 3d scene by primitives and gaussians.
\newblock \emph{arXiv preprint arXiv:2410.01535}, 2024.

\bibitem[Kerbl et~al.(2023)Kerbl, Kopanas, Leimk{\"u}hler, and Drettakis]{kerbl20233d}
Bernhard Kerbl, Georgios Kopanas, Thomas Leimk{\"u}hler, and George Drettakis.
\newblock 3d gaussian splatting for real-time radiance field rendering.
\newblock \emph{ACM Transactions on Graphics}, 42\penalty0 (4):\penalty0 1--14, 2023.

\bibitem[Kim et~al.(2024)Kim, Wu, Kerr, Tancik, Goldberg, and Kanazawa]{garfield2024}
Chung~Min* Kim, Mingxuan* Wu, Justin* Kerr, Matthew Tancik, Ken Goldberg, and Angjoo Kanazawa.
\newblock Garfield: Group anything with radiance fields.
\newblock In \emph{Conference on Computer Vision and Pattern Recognition (CVPR)}, 2024.

\bibitem[Kirillov et~al.(2023)Kirillov, Mintun, Ravi, Mao, Rolland, Gustafson, Xiao, Whitehead, Berg, Lo, Doll{\'a}r, and Girshick]{kirillov2023segany}
Alexander Kirillov, Eric Mintun, Nikhila Ravi, Hanzi Mao, Chloe Rolland, Laura Gustafson, Tete Xiao, Spencer Whitehead, Alexander~C. Berg, Wan-Yen Lo, Piotr Doll{\'a}r, and Ross Girshick.
\newblock Segment anything.
\newblock \emph{arXiv:2304.02643}, 2023.

\bibitem[Li et~al.(2017)Li, Xu, Chaudhuri, Yumer, Zhang, and Guibas]{li2017grass}
Jun Li, Kai Xu, Siddhartha Chaudhuri, Ersin Yumer, Hao Zhang, and Leonidas Guibas.
\newblock Grass: Generative recursive autoencoders for shape structures.
\newblock \emph{ACM Transactions on Graphics (TOG)}, 36\penalty0 (4):\penalty0 1--14, 2017.

\bibitem[Li et~al.(2019)Li, Sung, Dubrovina, Yi, and Guibas]{li2019supervised}
Lingxiao Li, Minhyuk Sung, Anastasia Dubrovina, Li Yi, and Leonidas~J Guibas.
\newblock {Supervised Fitting of Geometric Primitives to 3D Point Clouds}.
\newblock In \emph{CVPR}, 2019.

\bibitem[Li et~al.(2023)Li, M{\"u}ller, Evans, Taylor, Unberath, Liu, and Lin]{li2023neuralangelo}
Zhaoshuo Li, Thomas M{\"u}ller, Alex Evans, Russell~H Taylor, Mathias Unberath, Ming-Yu Liu, and Chen-Hsuan Lin.
\newblock Neuralangelo: High-fidelity neural surface reconstruction.
\newblock In \emph{Proceedings of the IEEE/CVF Conference on Computer Vision and Pattern Recognition}, pages 8456--8465, 2023.

\bibitem[Liu et~al.(2023)Liu, Yu, Ye, Zhangli, Li, Zhang, and Metaxas]{liu2023deformer}
Di Liu, Xiang Yu, Meng Ye, Qilong Zhangli, Zhuowei Li, Zhixing Zhang, and Dimitris~N Metaxas.
\newblock Deformer: Integrating transformers with deformable models for 3d shape abstraction from a single image.
\newblock In \emph{Proceedings of the IEEE/CVF International Conference on Computer Vision}, pages 14236--14246, 2023.

\bibitem[Liu et~al.(2022)Liu, Wu, Ruan, and Chirikjian]{liu2022robust}
Weixiao Liu, Yuwei Wu, Sipu Ruan, and Gregory~S Chirikjian.
\newblock {Robust and Accurate Superquadric Recovery: a Probabilistic Approach}.
\newblock In \emph{CVPR}, 2022.

\bibitem[Loiseau et~al.(2023)Loiseau, Vincent, Aubry, and Landrieu]{loiseau2023learnable}
Romain Loiseau, Elliot Vincent, Mathieu Aubry, and Loic Landrieu.
\newblock {Learnable Earth Parser: Discovering 3D Prototypes in Aerial Scans}.
\newblock \emph{arXiv:2304.09704 [cs.CV]}, 2023.

\bibitem[Mildenhall et~al.(2020)Mildenhall, Srinivasan, Tancik, Barron, Ramamoorthi, and Ng]{mildenhall2020nerf}
Ben Mildenhall, Pratul~P. Srinivasan, Matthew Tancik, Jonathan~T. Barron, Ravi Ramamoorthi, and Ren Ng.
\newblock {{NeRF}}: Representing {{Scenes}} as {{Neural Radiance Fields}} for {{View Synthesis}}.
\newblock In \emph{{{ECCV}}}, 2020.

\bibitem[Mitra et~al.(2013)Mitra, Wand, Zhang, Cohen-Or, Kim, and Huang]{Mitra_Wand_Zhang_Cohen-Or_Kim_Huang_2013}
Niloy Mitra, Michael Wand, Hao~(Richard) Zhang, Daniel Cohen-Or, Vladimir Kim, and Qi-Xing Huang.
\newblock Structure-aware shape processing.
\newblock In \emph{SIGGRAPH Asia 2013 Courses}, 2013.

\bibitem[Monnier et~al.(2023)Monnier, Austin, Kanazawa, Efros, and Aubry]{monnier2023dbw}
Tom Monnier, Jake Austin, Angjoo Kanazawa, Alexei~A. Efros, and Mathieu Aubry.
\newblock {Differentiable Blocks World: Qualitative 3D Decomposition by Rendering Primitives}.
\newblock In \emph{{NeurIPS}}, 2023.

\bibitem[Niu et~al.(2018)Niu, Li, and Xu]{Niu_Li_Xu_2018}
Chengjie Niu, Jun Li, and Kai Xu.
\newblock Im2struct: Recovering 3d shape structure from a single rgb image.
\newblock \emph{Cornell University - arXiv,Cornell University - arXiv}, 2018.

\bibitem[Paschalidou et~al.(2019)Paschalidou, Ulusoy, and Geiger]{paschalidou2019superquadrics}
Despoina Paschalidou, Ali~Osman Ulusoy, and Andreas Geiger.
\newblock Superquadrics {{Revisited}}: {{Learning 3D Shape Parsing Beyond Cuboids}}.
\newblock In \emph{{{CVPR}}}, 2019.

\bibitem[Paschalidou et~al.(2020)Paschalidou, Gool, and Geiger]{paschalidou2020learning}
Despoina Paschalidou, Luc~Van Gool, and Andreas Geiger.
\newblock {Learning Unsupervised Hierarchical Part Decomposition of 3D Objects from a Single RGB Image}.
\newblock In \emph{CVPR}, 2020.

\bibitem[Paschalidou et~al.(2021)Paschalidou, Katharopoulos, Geiger, and Fidler]{paschalidou2021neural}
Despoina Paschalidou, Angelos Katharopoulos, Andreas Geiger, and Sanja Fidler.
\newblock Neural parts: Learning expressive 3d shape abstractions with invertible neural networks.
\newblock In \emph{Proceedings of the IEEE/CVF Conference on Computer Vision and Pattern Recognition}, pages 3204--3215, 2021.

\bibitem[Porter and Duff(1984)]{porterCompositingDigitalImages1984}
Thomas Porter and Tom Duff.
\newblock Compositing {{Digital Images}}.
\newblock In \emph{{{SIGGRAPH}}}, 1984.

\bibitem[Ramamonjisoa et~al.(2022)Ramamonjisoa, Stekovic, and Lepetit]{ramamonjisoa2022monteboxfinder}
Micha{\"e}l Ramamonjisoa, Sinisa Stekovic, and Vincent Lepetit.
\newblock {MonteBoxFinder: Detecting and Filtering Primitives to Fit a Noisy Point Cloud}.
\newblock In \emph{{{ECCV}}}, 2022.

\bibitem[Roberts(1963)]{roberts1963machine}
Lawrence~G. Roberts.
\newblock \emph{Machine perception of three-dimensional solids}.
\newblock PhD thesis, Massachusetts Institute of Technology, 1963.

\bibitem[Sch\"{o}nberger and Frahm(2016)]{schoenberger2016sfm}
Johannes~Lutz Sch\"{o}nberger and Jan-Michael Frahm.
\newblock Structure-from-motion revisited.
\newblock In \emph{Conference on Computer Vision and Pattern Recognition (CVPR)}, 2016.

\bibitem[Schubert et~al.(2017)Schubert, Sander, Ester, Kriegel, and Xu]{schubert2017dbscan}
Erich Schubert, J{\"o}rg Sander, Martin Ester, Hans~Peter Kriegel, and Xiaowei Xu.
\newblock Dbscan revisited, revisited: why and how you should (still) use dbscan.
\newblock \emph{ACM Transactions on Database Systems (TODS)}, 42\penalty0 (3):\penalty0 1--21, 2017.

\bibitem[Shuai et~al.(2023)Shuai, Zhang, Yang, and Chen]{shuai2023dpf}
Qingyao Shuai, Chi Zhang, Kaizhi Yang, and Xuejin Chen.
\newblock Dpf-net: combining explicit shape priors in deformable primitive field for unsupervised structural reconstruction of 3d objects.
\newblock In \emph{Proceedings of the IEEE/CVF International Conference on Computer Vision}, pages 14321--14329, 2023.

\bibitem[Sun et~al.(2022)Sun, Yang, Guo, Wang, Tong, Liu, and Shum]{sun2022semi}
Chunyu Sun, Yuqi Yang, Haoxiang Guo, Pengshuai Wang, Xin Tong, Yang Liu, and Heung-Yeung Shum.
\newblock Semi-supervised 3d shape segmentation with multilevel consistency and part substitution.
\newblock \emph{Computational Visual Media}, 2022.

\bibitem[Tertikas et~al.(2023)Tertikas, Paschalidou, Pan, Park, Uy, Emiris, Avrithis, and Guibas]{tertikas2023partnerf}
Konstantinos Tertikas, Despoina Paschalidou, Boxiao Pan, Jeong~Joon Park, Mikaela~Angelina Uy, Ioannis Emiris, Yannis Avrithis, and Leonidas Guibas.
\newblock {{PartNeRF}}: {{Generating Part-Aware Editable 3D Shapes}} without {{3D Supervision}}.
\newblock In \emph{{{CVPR}}}, 2023.

\bibitem[Tulsiani et~al.(2017)Tulsiani, Su, Guibas, Efros, and Malik]{tulsiani2017learning}
Shubham Tulsiani, Hao Su, Leonidas~J. Guibas, Alexei~A. Efros, and Jitendra Malik.
\newblock Learning {{Shape Abstractions}} by {{Assembling Volumetric Primitives}}.
\newblock In \emph{{{CVPR}}}, 2017.

\bibitem[Waczy{\'n}ska et~al.(2024)Waczy{\'n}ska, Borycki, Tadeja, Tabor, and Spurek]{waczynska2024games}
Joanna Waczy{\'n}ska, Piotr Borycki, S{\l}awomir Tadeja, Jacek Tabor, and Przemys{\l}aw Spurek.
\newblock Games: Mesh-based adapting and modification of gaussian splatting.
\newblock \emph{arXiv preprint arXiv:2402.01459}, 2024.

\bibitem[Wan et~al.(2024{\natexlab{a}})Wan, Liu, Gan, Liu, Wang, Wen, Wan, and Zhu]{10486880}
Xinhang Wan, Jiyuan Liu, Xinbiao Gan, Xinwang Liu, Siwei Wang, Yi Wen, Tianjiao Wan, and En Zhu.
\newblock One-step multi-view clustering with diverse representation.
\newblock \emph{IEEE Transactions on Neural Networks and Learning Systems}, pages 1--13, 2024{\natexlab{a}}.

\bibitem[Wan et~al.(2024{\natexlab{b}})Wan, Liu, Yu, Qu, Li, Liu, Liang, Dong, and Zhu]{wan2024contrastive}
Xinhang Wan, Jiyuan Liu, Hao Yu, Qian Qu, Ao Li, Xinwang Liu, Ke Liang, Zhibin Dong, and En Zhu.
\newblock Contrastive continual multiview clustering with filtered structural fusion.
\newblock \emph{IEEE Transactions on Neural Networks and Learning Systems}, 2024{\natexlab{b}}.

\bibitem[Wang et~al.(2025{\natexlab{a}})Wang, Chen, Li, Wang, Wang, Guo, Wang, Shan, Lan, Wang, et~al.]{wang2025geollava}
Fengxiang Wang, Mingshuo Chen, Yueying Li, Di Wang, Haotian Wang, Zonghao Guo, Zefan Wang, Boqi Shan, Long Lan, Yulin Wang, et~al.
\newblock Geollava-8k: Scaling remote-sensing multimodal large language models to 8k resolution.
\newblock \emph{arXiv preprint arXiv:2505.21375}, 2025{\natexlab{a}}.

\bibitem[Wang et~al.(2025{\natexlab{b}})Wang, Wang, Guo, Wang, Wang, Chen, Ma, Lan, Yang, Zhang, et~al.]{wang2025xlrs}
Fengxiang Wang, Hongzhen Wang, Zonghao Guo, Di Wang, Yulin Wang, Mingshuo Chen, Qiang Ma, Long Lan, Wenjing Yang, Jing Zhang, et~al.
\newblock Xlrs-bench: Could your multimodal llms understand extremely large ultra-high-resolution remote sensing imagery?
\newblock In \emph{Proceedings of the Computer Vision and Pattern Recognition Conference}, pages 14325--14336, 2025{\natexlab{b}}.

\bibitem[Wang et~al.(2021)Wang, Liu, Liu, Theobalt, Komura, and Wang]{wang2021neus}
Peng Wang, Lingjie Liu, Yuan Liu, Christian Theobalt, Taku Komura, and Wenping Wang.
\newblock {{NeuS}}: {{Learning Neural Implicit Surfaces}} by {{Volume Rendering}} for {{Multi-view Reconstruction}}.
\newblock In \emph{{{NeurIPS}}}, 2021.

\bibitem[Wang et~al.(2020)Wang, Xu, Xu, Tagliasacchi, Zhou, Mahdavi-Amiri, and Zhang]{wang2020pie}
Xiaogang Wang, Yuelang Xu, Kai Xu, Andrea Tagliasacchi, Bin Zhou, Ali Mahdavi-Amiri, and Hao Zhang.
\newblock Pie-net: Parametric inference of point cloud edges.
\newblock \emph{Advances in neural information processing systems}, 33:\penalty0 20167--20178, 2020.

\bibitem[Wang et~al.(2004)Wang, Bovik, Sheikh, and Simoncelli]{wang2004image}
Zhou Wang, Alan~C Bovik, Hamid~R Sheikh, and Eero~P Simoncelli.
\newblock Image quality assessment: from error visibility to structural similarity.
\newblock \emph{IEEE transactions on image processing}, 13\penalty0 (4):\penalty0 600--612, 2004.

\bibitem[Wang et~al.(2025{\natexlab{c}})Wang, Yi, Wen, Zhu, and Xu]{Wang_2025_CVPR}
Zhifeng Wang, Renjiao Yi, Xin Wen, Chenyang Zhu, and Kai Xu.
\newblock Vastsd: Learning 3d vascular tree-state space diffusion model for angiography synthesis.
\newblock In \emph{Proceedings of the Computer Vision and Pattern Recognition Conference (CVPR)}, pages 15693--15702, 2025{\natexlab{c}}.

\bibitem[Wu et~al.(2021)Wu, Gao, Zhang, Lai, and Zhang]{STAR-TM}
Tong Wu, Lin Gao, Ling-Xiao Zhang, Yu-Kun Lai, and Hao Zhang.
\newblock Star-tm: Structure aware reconstruction of textured mesh from single image.
\newblock \emph{IEEE Transactions on Pattern Analysis and Machine Intelligence}, pages 1--14, 2021.

\bibitem[Wu et~al.(2022)Wu, Liu, Ruan, and Chirikjian]{wu2022primitivebased}
Yuwei Wu, Weixiao Liu, Sipu Ruan, and Gregory~S. Chirikjian.
\newblock Primitive-based {{Shape Abstraction}} via {{Nonparametric Bayesian Inference}}.
\newblock In \emph{{{ECCV}}}, 2022.

\bibitem[Yang et~al.(2022)Yang, Hu, Zhou, Liu, and Zhu]{ICL-SSL}
Xihong Yang, Xiaochang Hu, Sihang Zhou, Xinwang Liu, and En Zhu.
\newblock Interpolation-based contrastive learning for few-label semi-supervised learning.
\newblock \emph{IEEE Transactions on Neural Networks and Learning Systems}, 35\penalty0 (2):\penalty0 2054--2065, 2022.

\bibitem[Yang et~al.(2023)Yang, Liu, Zhou, Wang, Tu, Zheng, Liu, Fang, and Zhu]{CCGC}
Xihong Yang, Yue Liu, Sihang Zhou, Siwei Wang, Wenxuan Tu, Qun Zheng, Xinwang Liu, Liming Fang, and En Zhu.
\newblock Cluster-guided contrastive graph clustering network.
\newblock In \emph{Proceedings of the AAAI conference on artificial intelligence}, pages 10834--10842, 2023.

\bibitem[Yao et~al.(2020)Yao, Luo, Li, Zhang, Ren, Zhou, Fang, and Quan]{yao2020blendedmvs}
Yao Yao, Zixin Luo, Shiwei Li, Jingyang Zhang, Yufan Ren, Lei Zhou, Tian Fang, and Long Quan.
\newblock {{BlendedMVS}}: {{A Large-scale Dataset}} for {{Generalized Multi-view Stereo Networks}}.
\newblock In \emph{{{CVPR}}}, 2020.

\bibitem[Yariv et~al.(2020)Yariv, Kasten, Moran, Galun, Atzmon, Basri, and Lipman]{yariv2020multiview}
Lior Yariv, Yoni Kasten, Dror Moran, Meirav Galun, Matan Atzmon, Ronen Basri, and Yaron Lipman.
\newblock Multiview {{Neural Surface Reconstruction}} by {{Disentangling Geometry}} and {{Appearance}}.
\newblock In \emph{{{NeurIPS}}}, 2020.

\bibitem[Ye et~al.(2024)Ye, Danelljan, Yu, and Ke]{gaussian_grouping}
Mingqiao Ye, Martin Danelljan, Fisher Yu, and Lei Ke.
\newblock Gaussian grouping: Segment and edit anything in 3d scenes.
\newblock In \emph{ECCV}, 2024.

\bibitem[Ying et~al.(2023)Ying, Yin, Zhang, Wang, Yu, Huang, and Fang]{ying2023omniseg3d}
Haiyang Ying, Yixuan Yin, Jinzhi Zhang, Fan Wang, Tao Yu, Ruqi Huang, and Lu Fang.
\newblock Omniseg3d: Omniversal 3d segmentation via hierarchical contrastive learning.
\newblock \emph{arXiv preprint arXiv:2311.11666}, 2023.

\bibitem[Yu et~al.(2019)Yu, Liu, Zhang, Zhu, and Xu]{yu2019partnet}
Fenggen Yu, Kun Liu, Yan Zhang, Chenyang Zhu, and Kai Xu.
\newblock Partnet: A recursive part decomposition network for fine-grained and hierarchical shape segmentation.
\newblock In \emph{Proceedings of the IEEE/CVF conference on computer vision and pattern recognition}, pages 9491--9500, 2019.

\bibitem[Yu et~al.(2024)Yu, Qian, Zhang, Gil-Ureta, Jackson, Bennett, and Zhang]{yu2024dpa}
Fenggen Yu, Yimin Qian, Xu Zhang, Francisca Gil-Ureta, Brian Jackson, Eric Bennett, and Hao Zhang.
\newblock Dpa-net: Structured 3d abstraction from sparse views via differentiable primitive assembly.
\newblock \emph{arXiv preprint arXiv:2404.00875}, 2024.

\bibitem[Zhang et~al.(2018)Zhang, Isola, Efros, Shechtman, and Wang]{zhang2018unreasonable}
Richard Zhang, Phillip Isola, Alexei~A Efros, Eli Shechtman, and Oliver Wang.
\newblock The unreasonable effectiveness of deep features as a perceptual metric.
\newblock In \emph{Proceedings of the IEEE conference on computer vision and pattern recognition}, pages 586--595, 2018.

\bibitem[Zhou et~al.(2025)Zhou, Wang, Guo, and Xu]{zhou2025monomobilityzeroshot3dmobility}
Hongyi Zhou, Xiaogang Wang, Yulan Guo, and Kai Xu.
\newblock Monomobility: Zero-shot 3d mobility analysis from monocular videos, 2025.

\bibitem[Zhou et~al.(2018)Zhou, Park, and Koltun]{Zhou2018}
Qian-Yi Zhou, Jaesik Park, and Vladlen Koltun.
\newblock {Open3D}: {A} modern library for {3D} data processing.
\newblock \emph{arXiv:1801.09847}, 2018.

\bibitem[Zhu et~al.(2020)Zhu, Xu, Chaudhuri, Yi, Guibas, and Zhang]{zhu2020adacoseg}
Chenyang Zhu, Kai Xu, Siddhartha Chaudhuri, Li Yi, Leonidas~J Guibas, and Hao Zhang.
\newblock Adacoseg: Adaptive shape co-segmentation with group consistency loss.
\newblock In \emph{Proceedings of the IEEE/CVF Conference on Computer Vision and Pattern Recognition}, pages 8543--8552, 2020.

\end{thebibliography}
}
\clearpage
\setcounter{page}{1}
\maketitlesupplementary
At first, this supplementary material provides detailed experimental settings, including data processing procedures, implementation details for comparison baseline methods, and evaluation metrics (Sec.\ref{sec:exp_setting}). Subsequently, we provide further implementation specifics, including the training configurations and rendering process (Sec.\ref{sec:imp_detail}).  

\begin{figure*}[tbh]
  \centering
  \includegraphics[width=1.0\textwidth]{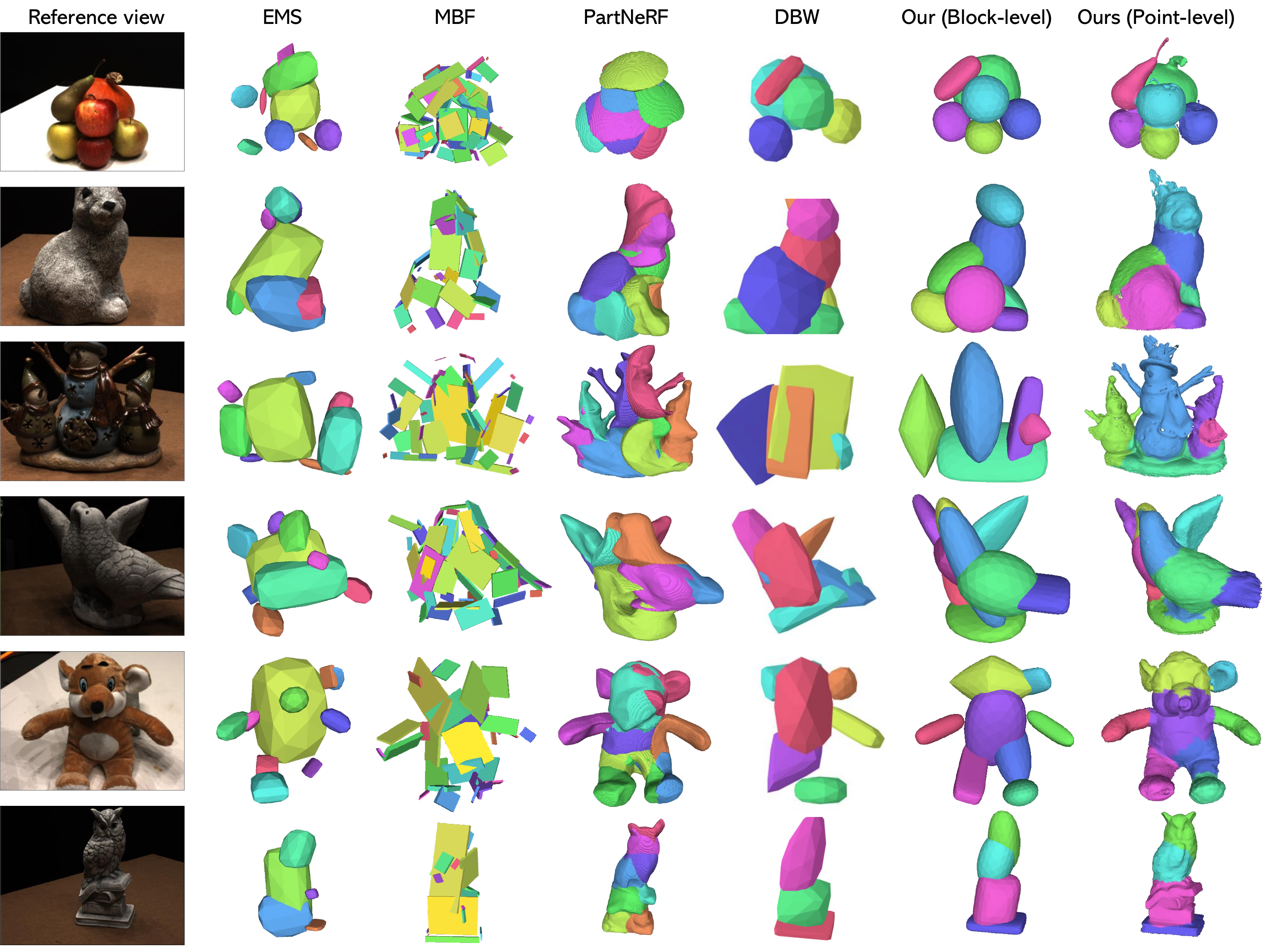}
  \caption{\textbf{Qualitative comparisons on DTU~\cite{jensen2014large}.} We compare our approach with state-of-the-art baselines on the DTU dataset with the background removed.  The last two columns show our block-level and point-level reconstructions, respectively.  Our method is the only one that provides reasonable 3D part decomposition while capturing detailed geometry.}
  \label{fig:dtu-part-supp}
\end{figure*}

\begin{figure*}[tbh]
  \centering
  \includegraphics[width=1.0\linewidth]{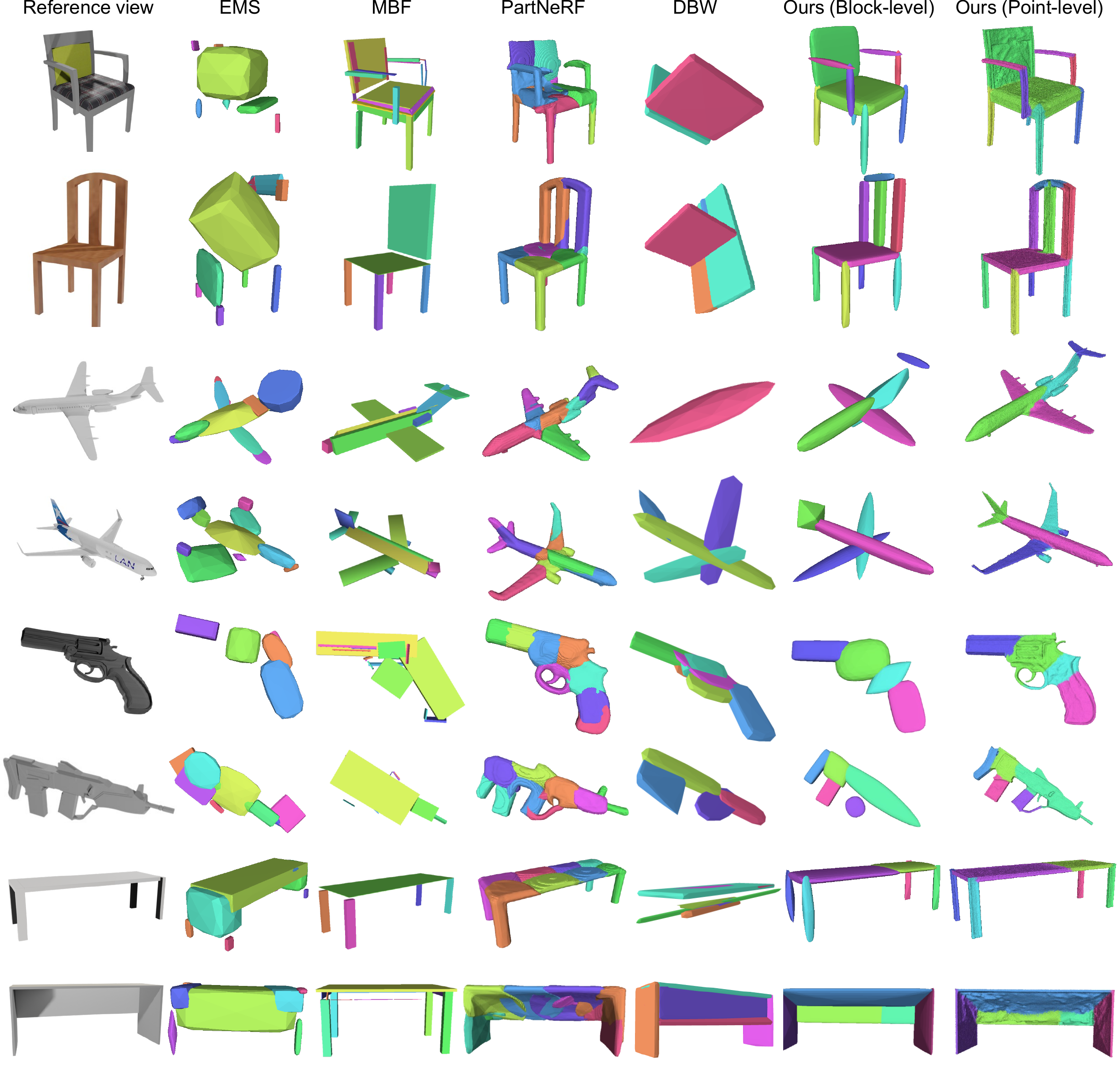}
  \caption{\textbf{Qualitative comparisons on ShapeNet~\cite{jensen2014large}.} We compared our approach with state-of-the-art baselines across four categories. The last two columns display our block-level and point-level reconstructions, respectively. Our method uniquely provides reasonable 3D part decomposition while simultaneously capturing detailed geometry.}
  \label{fig:shapent-part-supp}
\end{figure*}

In Fig.~\ref{fig:dtu-part-supp} and Fig.~\ref{fig:shapent-part-supp}, we provide additional visual comparisons of our method against state-of-the-art baselines on DTU~\cite{jensen2014large} and ShapeNet~\cite{chang2015shapenet} datasets. Moreover, we include visual results on several scenes from the BlendedMVS~\cite{yao2020blendedmvs} and self-captured data in Fig.~\ref{fig:mvs}. It can be observed that the decomposed reconstructions by our method are more reasonable and precise, which is believed to be beneficial for downstream tasks. Our code and data are available at \url{https://github.com/zhirui-gao/PartGS}.



\section{More Details on Experiments}
\label{sec:exp_setting}
\subsection{Datasets}
We conduct evaluations on two public datasets: DTU \cite{jensen2014large} and ShapeNet \cite{chang2015shapenet}. DTU is a multi-view stereo (MVS) dataset comprising 80 forward-facing scenes, each captured with 49 to 64 images.  In experiments, we use 15 publicly recognized scenes commonly adopted in previous studies~\cite{Huang2DGS2024, li2023neuralangelo}. We downsample the images to a resolution of $400\times300$ for computational efficiency.    

To validate the effectiveness of our method in the decomposition of man-made objects, we construct a subset of the ShapeNet dataset comprising four categories: \textit{Chair}, \textit{Table}, \textit{Gun}, and \textit{Airplane}. Each category includes 15 distinct objects, providing diverse instances. For each object, we randomly sample 100 camera poses on a sphere and render images at a resolution of $400\times400$. The rendered images are equally split into training and testing sets.

 Additionally, to explore the potential of PartGS in handling real-life data, we present qualitative results on the BlendedMVS dataset~\cite{yao2020blendedmvs} and self-captured scenes. For BlendedMVS, we use official camera poses, while for self-captured scenes, camera poses are estimated by COLMAP~\cite{schoenberger2016sfm} and normalized using IDR~\cite{yariv2020multiview}. All images are resized to $400\times300$ pixels.  For the
DTU, ShapeNet, and BlendedMVS datasets, we utilize the ground-truth foreground masks provided within the datasets. For self-captured real-world scenes, we employ the Segment
Anything Model (SAM)~\cite{kirillov2023segany} to segment foreground objects.

\subsection{Implementation Details on Baselines} 
We compare our method with four state-of-the-art works on 3D shape decomposition: EMS~\cite{liu2022robust},  MonteboxFinder (MBF)~\cite{ramamonjisoa2022monteboxfinder}, PartNeRF~\cite{tertikas2023partnerf}  and  DBW~\cite{monnier2023dbw}. MBF and EMS are applied to point clouds, utilizing cuboids and superquadrics, respectively, to fit the 3D points. The input point clouds are sampled from either ground truth 3D shapes (GT) or meshes reconstructed by state-of-the-art MVS method Neus~\cite{wang2021neus}. We adhere to the testing procedure outlined by DBW~\cite{monnier2023dbw} and randomly sample 5K and 200K points from the GT meshes as the input for EMS and MBF, respectively. Similar to our problem setting, PartNeRF and DBW use images as input to construct structural 3D representations. To ensure fair comparisons, we discard the pre-defined ground plane in DBW, as the test scenario consists exclusively of foreground objects. For PartNeRF, we perform instance-specific training and set the same number of parts as in our method, while retaining all other default parameter settings. Additionally, we compare with SOTA Gaussian Splatting reconstruction method 2DGS \cite{Huang2DGS2024}, and surface reconstruction approach Neuralangelo~\cite{li2023neuralangelo}, to provide an intuitive evaluation of the reconstruction quality achieved by our method.

\subsection{Evaluation Metrics} 
We evaluate from 3D reconstruction quality, view synthesis, shape parsimony, and reconstruction time. 

 \begin{itemize}

\item The 3D reconstruction quality is measured by the official Chamfer distance (CD) evaluation~\cite{jensen2014large} between the recovered geometry and GT, reflecting the accuracy of 3D reconstructions. 

\item View synthesis uses three standard metrics: Peak Signal-to-Noise Ratio (PSNR), Structural Similarity Index (SSIM)~\cite{wang2004image}, and Learned Perceptual Image Patch Similarity (LPIPS)~\cite{zhang2018unreasonable}. 

\item Shape parsimony is quantified by the number of parts, while reconstruction time refers to the running time measured on the same device. While shape parsimony does not directly indicate the quality of the decomposition, a smaller number of decomposed components, given consistent reconstruction quality, suggests a more concise and reasonable decomposition. This is because fitting different parts of an object with fewer blocks is inherently more challenging, whereas utilizing more blocks simplifies the fitting process.
\end{itemize}

\begin{figure*}[!ht]
  \centering
  \includegraphics[width=1.0\linewidth]{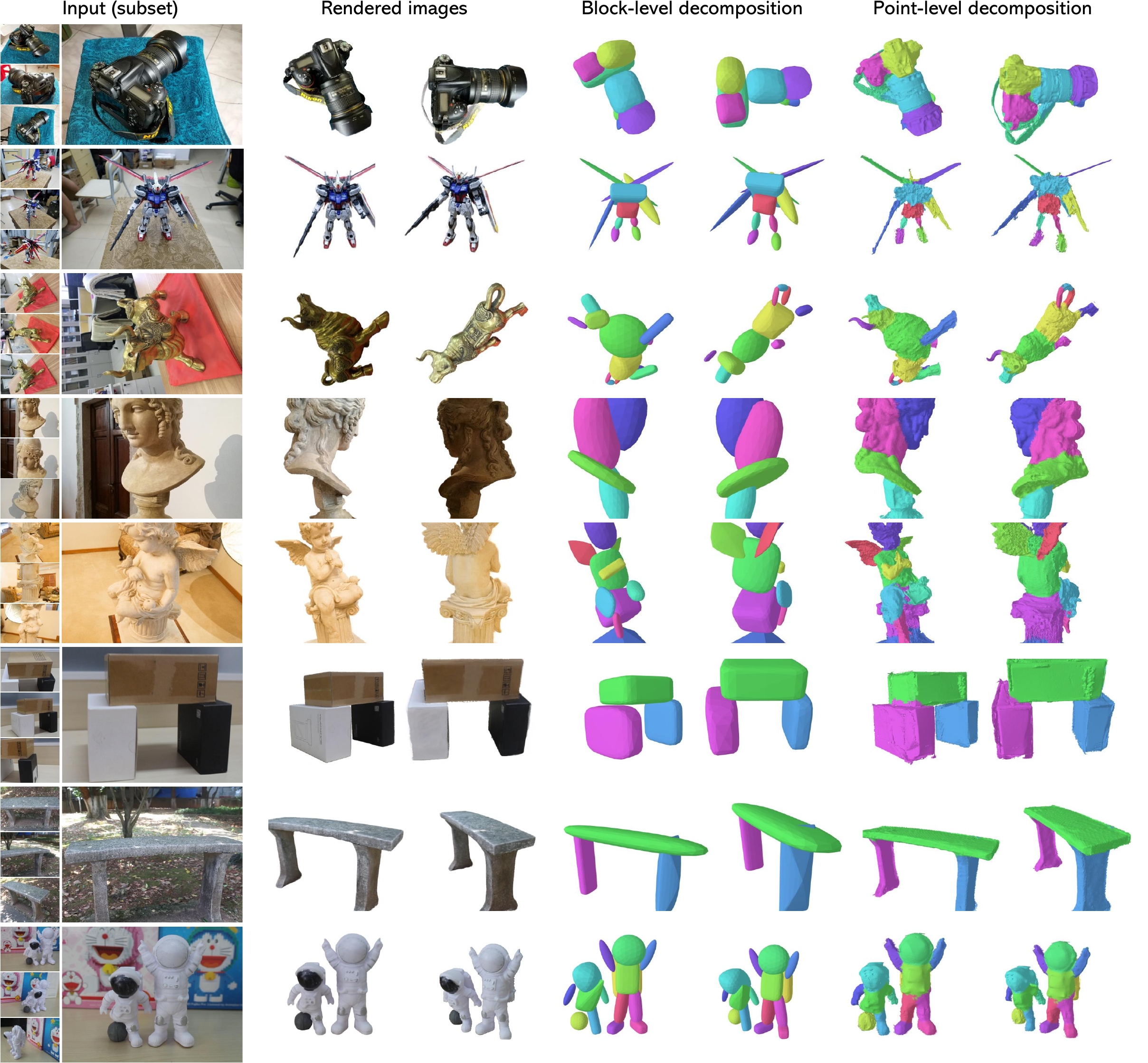}
  \caption{\textbf{Qualitative results on BlendedMVS~\cite{yao2020blendedmvs} and self-capatured data.}  We demonstrate the RGB  renderings and decomposed parts from novel views across a variety of objects. The first five examples are from the BlendedMVS dataset, and the remaining examples are from our own captured scenes.}
  \label{fig:mvs}
\end{figure*}

\section{Additional Comparative Experiments}

\begin{figure}[!h]
  \centering
  \includegraphics[width=\linewidth]{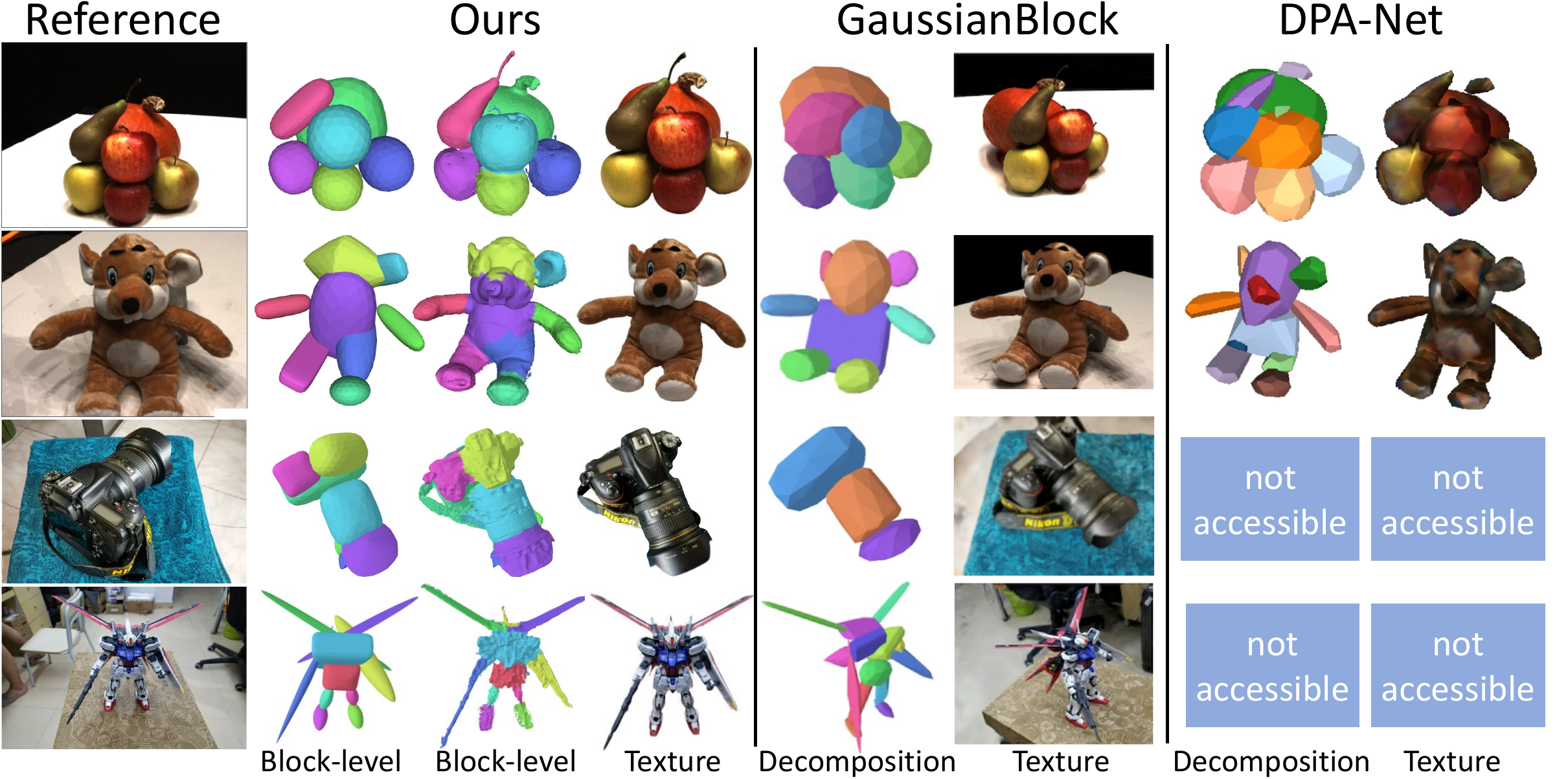}
  \caption{\textbf{Qualitative comparisons to DPA-Net and GaussianBlock.} The first two examples are from the DTU dataset, and the last two examples are from the BlendedMVS dataset. }
  \label{fig:fig1}
\end{figure}

Very recently, DPA-Net~\cite{yu2024dpa} and GaussianBlock~\cite{jiang2024gaussianblock} have achieved advanced 3D shape abstraction. DPA-Net enables part-aware reconstruction in a feedforward manner, requiring approximately 3 days of GPU training on a pre-collected dataset and 2 hours per object for inference. GaussianBlock utilizes SAM~\cite{kirillov2023segany} to guide superquadric splitting and fusion for 3D decomposition, with a processing time of 6 hours per object. We compare our method on DTU and BlendedMVS scenes, with results presented in Fig.~\ref{fig:fig1}. The proposed approach achieves comparable performance while demonstrating superior efficiency. It is important to clarify the supervision requirements: DPA-Net is a supervised method that relies on SAM-generated segmentation masks for training, whereas GaussianBlock, while not strictly 3D supervised, depends on high-quality pre-trained datasets that typically require days of training. Our method distinguishes itself by being completely self-supervised, requiring neither 3D supervision nor pre-trained components, yet achieving significantly better computational efficiency.

\section{Implementation Details} 
\label{sec:imp_detail}
\subsection{Training Configurations}
To ensure uniform distributions of 2D Gaussians across surfaces of the mesh, random barycentric coordinates are generated directly within each triangular face. Specifically,  barycentric weights are computed as $u=\sqrt{rand}$ and $v=rand$.  These weights are transformed to obtain $\alpha = [1 - u, u(1 - v), uv]$, $rand \in (0,1)$.  The sampled position on triangular face is calculated as $o = \alpha_0 v0 + \alpha_1 v1 + \alpha_2 v2 $, where $v_o, v_1,v_2$ are vertices of the triangle. This method compensates for the non-uniformity caused by the triangle’s geometry. Unlike naive random sampling which results in uneven distributions, it ensures that the sampled points are evenly distributed across mesh surfaces.

The same hyperparameters are used for all experiments. We set the initial number of primitives $M$ to 8. In the hybrid representation, each superquadric mesh is a level-2 icosphere (320 triangular face). Each triangular face contains 100 Gaussians, with a scaling parameter $c$ of 0.1. The number of sampled points in each ray is 2048.   

During refinement, we employ regularization in 2DGS to achieve better geometric reconstruction, including depth distortion maps, depth maps, and normal maps. Additionally, we introduce a mask cross-entropy to filter out extra noise Gaussians. To extract the meshes from 2D Gaussians, we use truncated signed distance fusion (TSDF) to fuse rendered depth maps, utilizing Open3D~\cite{Zhou2018}.

\subsection{Rendering}
With Gaussians attached to the surface of each block, we achieve view-dependent rendering through tile-based rasterization as Gaussian Splatting \cite{kerbl20233d, Huang2DGS2024}.  Given a view, Gaussians on the superquadric surface are projected onto the image space, forming an RGB image.  Initially, the screen space determines the bounding box for each Gaussian.  Subsequently,  these Gaussian ellipses are sorted according to their depths of center to the image plane. Finally, volumetric alpha compositing \cite{porterCompositingDigitalImages1984} is utilized to integrate the alpha-weighted RGB values for each pixel.

To formulate the process, considering a viewing transformation ${W}$, the covariance matrix ${\Sigma}_{i}^{'}$ of $i$-th Gaussian in the camera coordinate system is calculated by:
\begin{equation}
   {\Sigma}_{i}^{'} = {J}{W}{\Sigma}{W}^{T}{J}^{T},
\end{equation}
where ${J}$ is the Jacobian of the affine approximation of the projective transformation, and ${\Sigma} $ is the covariance matrix of the Gaussian ellipse. Note that the last row and column of ${\Sigma}$ are omitted since we adopt 2D Gaussians.
Following alpha compositing, we first calculate an alpha value for each Gaussian ellipse:
\begin{equation}
    \alpha_i = \tau_i\exp(-\frac{1}{2}(\mathbf{x}-\mathbf{u}_{i})^{T}{\Sigma_{i}^{'}}_{-1}(\mathbf{x}-\mathbf{u}_{i}))).
\end{equation}
Here, $\mathbf{u}_{i}$ is the center coordinate of the projected Gaussian ellipse, and $\tau_i$ is the opacity of the block where the $i$-th Gaussian is located. The calculated alphas are sorted according to their depths from the image plane. Meanwhile, we can acquire the color value $c_i$ from the spherical harmonics of $\mathcal{N}$ ordered points, thus obtaining the rendering RGB value:
\begin{equation}
    C = \sum_{i\in \mathcal{N}}c_i\alpha_i\prod_{j=1}^{i-1}(1-\alpha_i).
    \label{eq4}
\end{equation}
This process is differentiable and can optimize the hybrid representation through gradient descent.

\end{document}